\documentclass[preprint,12pt]{elsarticle}

\usepackage{amssymb}
\usepackage{subcaption}
\usepackage{multirow}
\usepackage{algorithm2e}
\RestyleAlgo{ruled}
\usepackage{longtable}
\usepackage{amsmath}
\usepackage{natbib}
\usepackage{booktabs}

\journal{Applied Soft Computing}

\begin{document}

\begin{frontmatter}



\title{Twitter conversations predict the daily confirmed COVID-19 cases}

\author[inst1]{Rabindra Lamsal\corref{cor1}}
\author[inst1]{Aaron Harwood}
\author[inst1]{Maria Rodriguez Read}

\affiliation[inst1]{organization={School of Computing and Information Systems, \\ The University of Melbourne},
            addressline={Parkville}, 
            city={Melbourne},
            postcode={3010}, 
            state={Victoria},
            country={Australia}}

\cortext[cor1]{rlamsal@student.unimelb.edu.au}

\begin{abstract}
As of writing this paper, COVID-19 (Coronavirus disease 2019) has spread to more than 220 countries and territories. Following the outbreak, the pandemic’s seriousness has made people more active on social media, especially on the microblogging platforms such as Twitter and Weibo. The pandemic-specific discourse has remained on-trend on these platforms for months now. Previous studies have confirmed the contributions of such socially generated conversations towards situational awareness of crisis events. The early forecasts of cases are essential to authorities to estimate the requirements of resources needed to cope with the outgrowths of the virus. Therefore, this study attempts to incorporate the public discourse in the design of forecasting models particularly targeted for the steep-hill region of an ongoing wave. We propose a sentiment-involved topic-based latent variables search methodology for designing forecasting models from publicly available Twitter conversations. As a use case, we implement the proposed methodology on Australian COVID-19 daily cases and Twitter conversations generated within the country. Experimental results: (i) show the presence of latent social media variables that Granger-cause the daily COVID-19 confirmed cases, and (ii) confirm that those variables offer additional prediction capability to forecasting models. Further, the results show that the inclusion of social media variables introduces 48.83--51.38\% improvements on RMSE over the baseline models. We also release the large-scale COVID-19 specific geotagged global tweets dataset, \textit{MegaGeoCOV}, to the public anticipating that the geotagged data of this scale would aid in understanding the conversational dynamics of the pandemic through other spatial and temporal contexts.
\end{abstract}




\end{frontmatter}



\section{Introduction}
COVID-19 (Coronavirus disease 2019) is a respiratory illness caused by severe acute respiratory syndrome coronavirus 2 (SARS-CoV-2), the first case identified in Wuhan, China, in December 2019, has since spread globally, spanning to an ongoing pandemic \cite{zhang2020clinical}. The disease was declared as a public health emergency of international concern on January 30, 2020, and as a pandemic on March 11, 2020, by the World Health Organization. As of November 9, 2021, more than 251 million global cases and more than 5 million deaths have been confirmed \cite{worldometers}. 
During the early phase of the pandemic, countries and territories around the globe initiated partial and/or complete lockdowns to contain the spread of the virus. Mass vaccination campaigns have also been started after late 2020 with vaccines such as Oxford-AstraZeneca, Pfizer, Moderna, Johnson and Johnson, and Sinovac \cite{approvedvaccines}. 

\begin{figure}[h]
  \includegraphics[width=\linewidth]{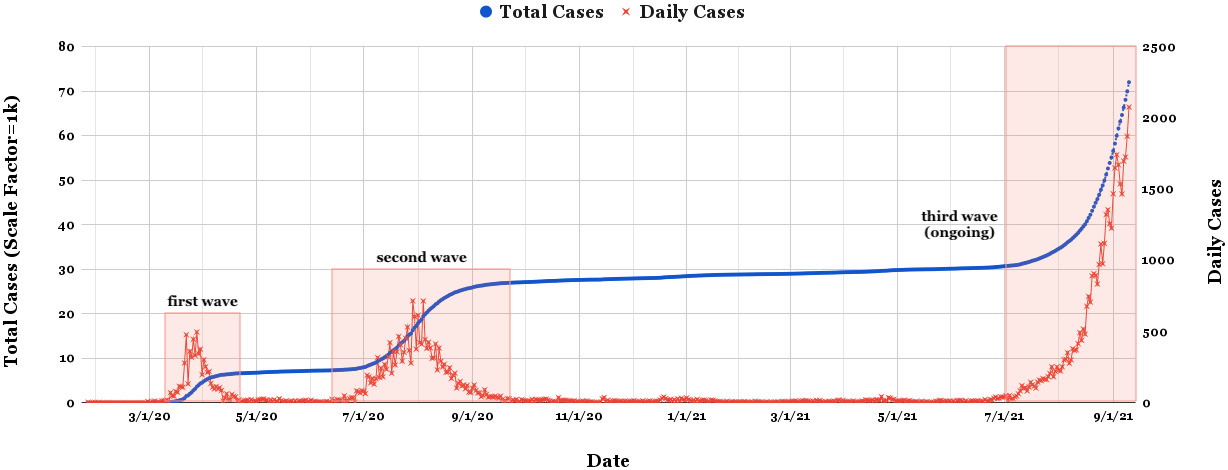}
  \caption{Daily (new) and total (cumulative) COVID-19 cases reported in Australia between January 25, 2020 (first COVID-19 case reported), and September 9, 2020.}
  \label{covid-cases}
\end{figure}

In the case of Australia, the country's first case of COVID-19 was confirmed by Victoria Health Authorities on January 25, 2020 \cite{first_aus_case}. Since then, as of November 9, 2021, 182,870 cases and 1,841 deaths have been confirmed as the country is currently facing its third wave of COVID-19 infections \cite{worldometers}. Figure \ref{covid-cases} shows the daily confirmed numbers and the cumulative numbers of COVID-19 infections in Australia between late January 2020 and early September 2021. Also illustrated in Figure \ref{covid-cases}, Australia experienced the first wave of COVID-19 infections during March--April 2020, the second wave during June--October 2020, and the ongoing third wave since June 2021. Other than during the waves, the daily COVID-19 infections in Australia are within two digits. The highest confirmed cases for a single day during the first wave were 497, during the second wave were 716; while the third wave is ongoing and reporting significantly large figures each day \cite{worldometers}.

Since the outbreak, the pandemic's gravity has made people more vocal on social media, especially on microblogging platforms such as Twitter and Weibo. As people share what they are experiencing, observing, and gathering, multiple terms related to the pandemic have emerged and remained on-trend on these platforms for months now. Previous studies have shown that such public discourse contributes to a better understanding of an ongoing crisis. With this consideration, this study attempts to incorporate the public discourse in designing pandemic-related time series forecasting models \textit{specially targeted for the steep-hill region of a pandemic's ongoing wave}. The modeling and early prediction of the prevalence of virus are essential to provide situational information to decision making bodies and authorities to estimate the requirements of resources and equipment needed to cope with the consequences of the virus \cite{lamsal2022socially}. This study, therefore, focuses on the forecast of COVID-19 spread while addressing the following research questions (RQs):

\textbf{RQ1}: Geotagged data plays a crucial role while modeling location-specific information \cite{comito2021covid}. Inclusion of social media variables into forecast models requires a large amount of geotagged data. Therefore, we would like to know what portion of the Twitter volume is geotagged? After the release of Twitter's Academic Track-based Full-archive search and count APIs, finally, it is possible to address this research question---earlier researchers were able to access only a sample of overall Twitter volume. 

\textbf{RQ2}: Is there a presence of latent variables within geotagged Twitter data that Granger-cause the daily COVID-19 confirmed cases time series?

\textbf{RQ3}: If the answer to \textbf{RQ2} is `yes', do those variables provide additional prediction capability to time series forecasting models?

\textbf{RQ4}: Is ``the volume of public discourse in the last few days" predictive of the steep-hill curve of COVID-19 cases during an ongoing wave?

The paper is organized as follows: Section \ref{relatedwork} presents related work, Section \ref{timeseriesdataset} explains the design of the time series dataset (includes data collection, sentiment analysis, topic modelling), Section \ref{experimentation} presents experimentation and discussion, and Section \ref{conclusion} is the conclusion.

\section{Related Work}
\label{relatedwork}

In the past, modeling and forecasting of cases and transmission risks have been done across multiple areas: human West Nile virus cases and mosquito infection rates \cite{defelice2017ensemble}, hepatitis A virus infection \cite{ture2006comparison}, seasonal outbreaks of influenza \cite{shaman2012forecasting} and its real-time tracking \cite{shaman2013real}, Ebola outbreak \cite{shaman2014inference}, H1N1-2009 \cite{ong2010real}, international spread of Middle East respiratory syndrome (MERS) \cite{nah2016predicting}. There have also been some notable works in the area of forecasting the daily confirmed cases of the ongoing pandemic. Maleki et al. \cite{maleki2020time} modeled the total number of global confirmed cases and recovered cases using autoregressive models based on the two-piece $t$ distributions for predicting the global cases between April 21, 2020--April 31, 2020. In \cite{salgotra2020time}, Salgotra et al. performed time series prediction of COVID-19 confirmed and death cases across major Indian cities for the period May 15, 2020--May 25, 2020, based on genetic programming \cite{koza1992genetic}. Papastefanopoulos et al. \cite{papastefanopoulos2020covid} used both traditional statistical methods and machine learning approaches for estimating the percentage of active cases per population, up to 7 days into the future, for ten countries including the United States, Spain, Italy, the United Kingdom, and Germany.  The authors showed that, overall, the traditional approaches like ARIMA (Autoregressive Integrated Moving Average) prevail over methods based on machine learning in the forecast of COVID-19 time series due to lack of a large amount of data. Similarly, Saba et al. \cite{saba2021machine} observed ARIMA and SARIMA (Seasonal ARIMA) models producing relatively better results, in the forecast of daily COVID-19 cases during complete and herd lockdowns, than machine learning algorithms such as Polynomial Regression, K-nearest neighbors, Random Forests, Support Vector Machine, and Decision Trees. In \cite{singh2020development}, Singh et al. used a hybrid model with discrete wavelet decomposition and ARIMA to forecast the cases of COVID-19.

ARIMA and its variations appear to be the most favored techniques for COVID-19 cases time series forecast. Different parameterized ARIMA and its variants have been used across studies targeting regions such as India \cite{khan2020arima, sahai2020arima}, Pakistan \cite{yousaf2020statistical}, Saudi Arabia \cite{alzahrani2020forecasting}, Mainland China, Italy, South Korea, Thailand \cite{dehesh2020forecasting}, US, Brazil, Russia, Spain \cite{sahai2020arima}, North America, South America, Africa, Asia and Europe \cite{hernandez2020forecasting}, Italy, Spain, and France \cite{ceylan2020estimation}, and the most hit countries \cite{singh2020prediction, arunkumar2021forecasting}. Furthermore, models such as Susceptible, Exposed, Infection and Recover (SEIR), Infection and Recover (SIR) and their variations, and others such as Agent-based models, Curve-fitting and Logistic growth models have also been applied extensively for mathematical modeling of the COVID-19 situations for forecasting purposes \cite{malavika2021forecasting, arti2020modeling, pandey2020seir, ivorra2020mathematical, mandal2020prudent, huang2020epidemic}.

Social media platforms, such as Facebook and Twitter, have an active user base of millions and hold an enormous amount of socially generated data through the exchange of conversations. These platforms have become an active source of information during day-to-day life as well as during mass emergencies such as the ongoing pandemic \cite{lamsal2022socially}. During mass emergencies, the number of user activities across these platforms increases exponentially, as people: (a) generate trends on search engines such as Google, (b) share their safety status or query the safety status of their near ones, and (c) also share what they have seen, felt, or heard. These socially generated activities can be collected and analyzed for understanding the relationship between public discourse and how an emergency event unfolds at the ground level \cite{lamsal2022socially}. For example, in \cite{chew2021hybrid}, Chew et al. used semantic word vectors as a representation of the public's response to the pandemic to forecast the daily growth rate in the number of global confirmed COVID-19 cases with a lead-time of 1 day for the period January 25, 2020, and May 11, 2020. The authors extracted vector representations from more than 100 million English language tweets, trained a deep neural network on the vectors alongside the historical time series of growth rates and reported that their neural nets based approach outperforms traditional time series and machine learning models. In \cite{qin2020prediction} Qin et al. collected social media search indexes (SMSI) for the COVID-19 specific keywords---dry cough, chest distress, coronavirus, fever, and pneumonia---from December 31, 2019, to February 9, 2020. The authors used the lagged series of the search indexes to predict the COVID-19 case numbers for the same period and reported that the cases' trend correlated significantly with the lagged series. COVID-19 specific search query volumes on Google, Baidu, and Weibo have also been observed correlated to laboratory-confirmed and suspected cases of COVID-19 \cite{li2020retrospective}. Similar results were reported by Cousins et al. \cite{cousins2020regional}---the search-engine query patterns were observed predictive of COVID-19 case rates.

In \cite{li2020data}, Li et al. collected around 115k Weibo posts originating from Wuhan, China, between December 23, 2019, and January 30, 2020, and designed a regression model to observe the COVID-19 related posts being predictive of the number of cases reported. Similarly, Shen et al. \cite{shen2020using} used more than 15 million Weibo posts created between November 1, 2019, and March 31, 2020, and designed a machine learning classifier to identify ``sick" related posts. The count of such ``sick" posts were observed Granger-causing the daily number of COVID-19 cases. In \cite{comito2021covid}, Comito reported that the number of Twitter posts increases before confirmed cases follow a similar trend, suggesting that social media discourse can be a front indicator of epidemics spreading.

The studies dealing with the early forecasts of confirmed cases, may it be COVID-19 or previous epidemics outbreaks, rely majorly on the ``volume of conversations" feature, i.e. overall count, sentiment-based count, or a specific category-based count \cite{lamsal2022socially}. The issue with ``volume of conversations" feature is its reliability and robustness. Methodologies based on this feature get heavily affected by avalanches of autogenerated conversations. Furthermore, as per our literature search, the effectiveness of the latent variables within the publicly available social media conversations has not been studied for their possible influence on the trend of a pandemic/epidemic outbreak. While addressing these limitations, this study contributes the following to the literature:\\

\noindent\textbf{(a)} the study proposes an effective representation for microblog conversations, such that the ``volume of conversations" feature can be represented at a more granular level to decrease the intensity of possible forecast biases,\\
\textbf{(b)} the study provides evidences that confirm the significance of social media variables in forecasting the future trend of a steep-hill curve of a pandemic/epidemic outbreak, and\\
\textbf{(c)} we release a large-scale COVID-19 specific geotagged tweets dataset, MegaGeoCOV\footnote{https://github.com/rabindralamsal/MegaGeoCOV}, to the public. The dataset was curated for this study, and as per Twitter's terms of use \cite{twitter_dev_tos}, we only release the tweet identifiers, which can be hydrated using tools such as Hydrator\footnote{https://github.com/DocNow/hydrator} (desktop application) or twarc\footnote{https://twarc-project.readthedocs.io/en/latest/} (python library) to rebuild the dataset locally.


\section{Time series}
\label{timeseriesdataset}

\begin{figure}[h!]
  \includegraphics[width=\linewidth]{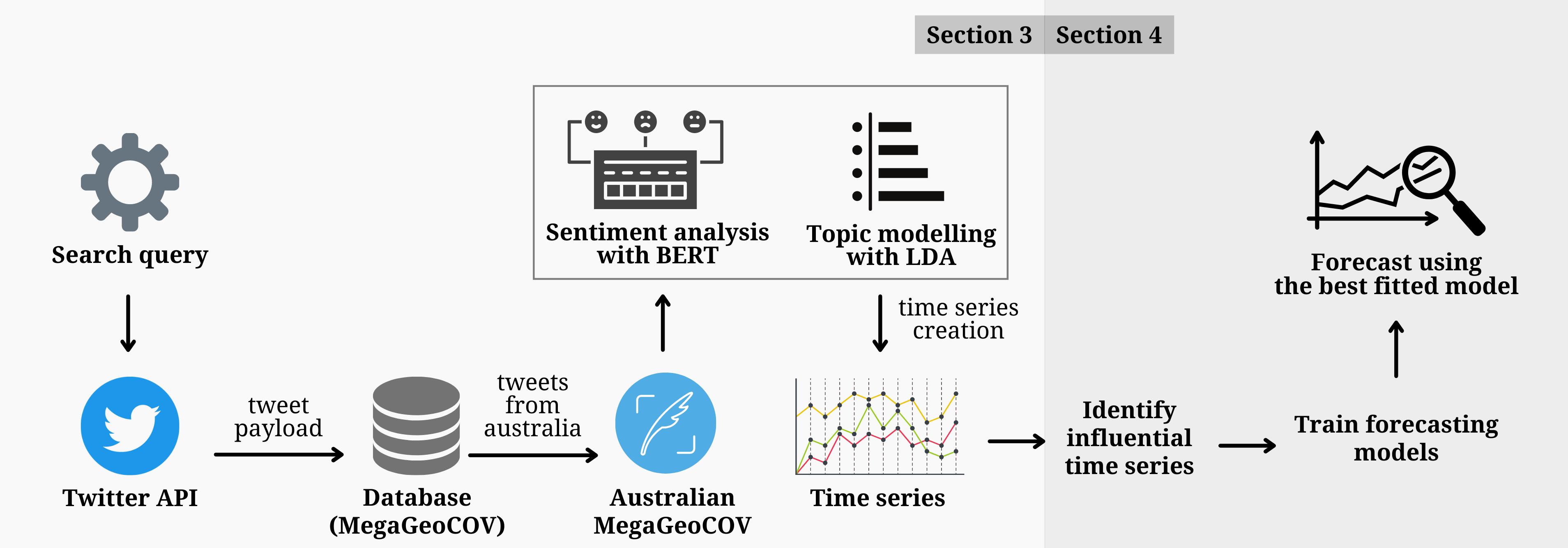}
  \caption{The overall view of the Twitter-based COVID-19 cases forecast methodology.}
  \label{process}
\end{figure}

We implement the methodology illustrated in Figure \ref{process} for our time series analysis. In this section, we discuss the data collection procedure and the time series formulation approach in detail, and in the next section (Section \ref{experimentation}), we design the forecasting models on a set of influential time series and experiment with social media variables in the design of pandemic related forecasting models to address our research questions.

\subsection{Data Collection}
\label{datacollection}
We considered Twitter as a primary data source since its acts as an instant, short, and frequent basis of communication, and most importantly it allows researchers to access the publicly available data on its platform through a wide range of API endpoints \cite{twitter_apis}. Some of the widely used Twitter's endpoints include \textit{Tweet lookup endpoint} for looking up tweets using tweet identifiers, \textit{Search endpoint} for searching most recent 7 days, or the full-archive of tweets, \textit{Tweet counts endpoint} for retrieving a count of tweets that match given query, \textit{Filtered stream endpoint} for retrieving real-time public tweets, and \textit{Sampled stream endpoint} for retrieving approximately 1\% of all real-time public Tweets.

In this study, we used Twitter's new academic track endpoint, the Full-archive search endpoint\footnote{This new endpoint enables researchers to collect tweets from as early as 2006.}, for collecting COVID-19 specific tweets created between January 01, 2020, and September 9, 2021. The following keywords (plain word) and hashtags (word preceeded by \texttt{\#} symbol) were considered while searching and collecting for COVID-19 specific tweets: \texttt{coronavirus}, \texttt{\#coronavirus}, \texttt{covid}, \texttt{\#covid}, \texttt{covid19}, \texttt{\#covid19}, \texttt{covid-19}, \texttt{\#covid-19}, \texttt{pandemic}, \texttt{\#pandemic}, \texttt{quarantine}, \texttt{\#quarantine}, \texttt{\#lockdown}, \texttt{lockdown}, \texttt{ppe}, \texttt{n95}, \texttt{\#ppe}, \texttt{\#n95}, \texttt{pneumonia}, \texttt{\#pneumonia}, \texttt{virus}, \texttt{\#virus}, \texttt{mask}, \texttt{\#mask}, \texttt{vaccine}, \texttt{vaccines}, \texttt{\#vaccine}, \texttt{\#vaccines}, \texttt{lungs}, and \texttt{flu}. The keywords selection was done based on previously proposed sets of keywords \cite{lamsal2021design, 781w-ef42-20}. Additionally, we used the Full-archive count API for getting the descriptive statistics (presented in Table \ref{stat}) of the daily COVID-19 public discourse on Twitter.

\begin{figure}[h]
  \includegraphics[width=\linewidth]{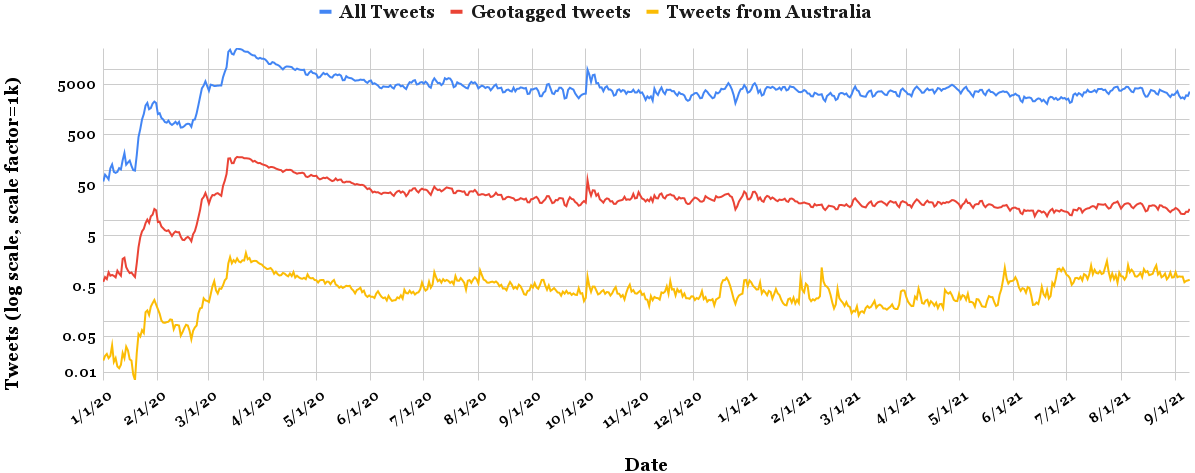}
  \caption{Daily distribution of COVID-19 specific tweets between January 1, 2020, and September 9, 2020.}
  \label{tweets-count}
\end{figure}

Generally, there are two classes of geographical metadata available with tweets. The first class is related to ``tweet location" in which a location is shared by a Twitter user while creating a tweet. The location data is attached with the tweet either as exact geocoordinates (a point location) or as a bounding box (a general location). The second class is related to ``account location" which is based on the location provided by a user on his/her public profile. Since the account location field is not validated by Twitter, we only considered the tweets having exact geocoordinates or bounding boxes while designing the forecasting models. In total, 21.36 million geotagged COVID-19 specific tweets were retrieved from the API endpoint. We name this large-scale geotagged global tweets dataset \textit{MegaGeoCOV}. The dataset is briefly explored in terms of numbers across multiple attributes (general overview given in Table \ref{MegaGeoCOV})---countries, cities and states (in Table \ref{MegaGeoCOV-location}), languages (in Table \ref{MegaGeoCOV-language}), and frequency distribution (in Figure \ref{tweets-count}). 

\begin{table}[!h]
\footnotesize
\centering
\caption{Descriptive statistics of the daily COVID-19 public discourse on Twitter.}
\label{stat}
\begin{tabular}{p{2cm}|l|p{2cm}|p{2cm}|p{2cm}}
\hline
& \textbf{All Tweets} & \textbf{Geotagged Tweets} (global) & \textbf{Tweets from} \newline \textbf{Australia} & \textbf{\% of tweets geotagged} (global)\\
\hline
\textbf{mean} & 4.62 million & 33.2k & 493 & 0.694538\\
\hline
\textbf{std} & 3.74 million & 30.8k & 337 & 0.129555\\
\hline
\textbf{minimum} & 59.6k & 615 & 7 & 0.449497\\
\hline
\textbf{25\%} & 3.06 million & 18.9k & 272 & 0.595030
\\
\hline
\textbf{median} & 3.77 million & 24.4k & 408 & 0.682665
\\
\hline
\textbf{75\%} & 4.64 million & 33.7k & 660 & 0.781970\\
\hline
\textbf{maximum} & 25.8 million & 183k & 2297 & 1.439504\\
\hline
\end{tabular}
\end{table}

\begin{table}[!h]
\footnotesize
\centering
\caption{Overview of MegaGeoCOV.}
\label{MegaGeoCOV}
\begin{tabular}{p{6cm}|p{3cm}}
\hline
\textbf{Total tweets (unique ids)} & 21,305,691\\
\hline
\textbf{Duplicate tweets (exact copy)} & 137,836\\
\hline
\textbf{Countries and territories} & 245\\
\hline
\textbf{Cities and states} & 260,732\\
\hline
\textbf{Languages} & 64 (and undefined)\\
\hline
\end{tabular}
\end{table}

\begin{table}[!htb]
\footnotesize
    \caption{Top 15 global locations in MegaGeoCOV.}
    \label{MegaGeoCOV-location}
    \begin{subtable}{.5\linewidth}
      \centering
        \caption{Top countries/territories (N=245)}
        \begin{tabular}{p{3.2cm}|p{2cm}}
        \hline
        \textbf{Country/territory} & \textbf{Tweets}\\
        \hline
        United States & 7,375,997\\
        United Kingdom & 2,279,064\\
        India & 1,563,017\\
        Brazil & 1,379,733\\
        Canada & 756,466\\
        Spain & 625,599\\
        Indonesia & 509,498\\
        Argentina & 434,454\\
        Mexico & 430,478\\
        Philippines & 383,215\\
        Australia & 366,033\\
        South Africa & 357,674\\
        France & 339,001\\
        Italy & 324,028\\
        Nigeria & 293,242\\
        \hline
        \end{tabular}
    \end{subtable}%
    \begin{subtable}{.5\linewidth}
      \centering
        \caption{Top cities and states (N=260,732)}
        \begin{tabular}{p{3cm}|p{2cm}}
        \hline
        \textbf{City/state} & \textbf{Tweets}\\
        \hline
        Los Angeles & 240,374\\
        Rio de Janerio & 192,986\\
        Manhattan & 185,021\\
        New Delhi & 173,854\\
        Mumbai & 155,855\\
        Sao Paulo & 148,202\\
        Toronto & 141,963\\
        Florida & 122,370\\
        Chicago & 120,930\\
        Brooklyn & 112,231\\
        Houston & 111,836\\
        Melbourne & 111,038\\
        Washington & 98,907\\
        Madrid & 96,592\\
        Buenos Aires & 95,759\\
        \hline
        \end{tabular}
    \end{subtable}
\end{table}

\begin{table}[!h]
\footnotesize
\centering
\caption{Most frequent languages (N=64) in MegaGeoCOV.}
\label{MegaGeoCOV-language}
\begin{tabular}{@{}p{4cm}|l|p{3cm}@{}}
\hline
\textbf{Language} & \textbf{ISO}$^a$ & \textbf{Tweets}\\
\hline
English & en & 13,854,642\\
Spanish & es & 2,545,726\\
Portuguese & pt & 1,389,951\\
Indonesian & in & 708,023\\
Undefined & - & 689,301\\
French & fr & 415,434\\
Italian & it & 280,087\\
Tagalog & tl & 274,845\\
Hindi & hi & 221,280\\
Turkish & tr & 157,962\\
German & de & 143,874\\
\hline
\multicolumn{3}{l}{Other languages$^a$ in order of their frequencies:}\\
\multicolumn{3}{l}{\begin{tabular}[c]{@{}l@{}}nl, ca, ja, th, ar, pl, et, ru, sv, ht, lt, mr, ro, cs, fi, da, el, ur, ta, \\ zh, sl, ne, gu, bn, lv, no, vi, cy, te, kn,  uk, hu, ko, or, fa, is, eu, \\ si, ml, iw, bg, sr, pa, dv, km, my, am, sd, ckb, ps, lo, hy, ka, bo\end{tabular}} \\
\hline
\end{tabular}
\\ $^a$ISO 639-1 Language Code
\end{table}

\subsubsection{Australian Tweets}
\textit{MegaGeoCOV} has more than 90 tweet objects, each object representing various tweet metadata. From \textit{MegaGeoCOV}, we extracted tweets originating from Australia (by conditioning the \texttt{geo.country} object) and considered only the \texttt{created\_at} (date and time), \texttt{text} (tweet), \texttt{geo.full\_name} (geolocation) objects for curating Australia-specific COVID-19 tweets dataset, from here on termed as dataset $D$. Since the \texttt{geo.full\_name} object followed the [city, state] data structure, all other geolocation-specific objects were ignored as this object was enough for extracting both city- and state-level information.

\textbf{Tweets selection.} Out of the $366,033$ tweets originating from Australia, only the tweets geotagged with exact geocoordinates or bounding box coordinates were considered. Twitter does not validate the account location field. Entries such as ``My Home", ``My Dream", ``Solar System", ``Milky Way Galaxy", etc are equally valid. Further, some users can have one location on their public profile and create tweets from some other location. Therefore we considered only the tweets whose geolocation was shared by users while creating tweets. Next, we filtered out tweets that had less than $10$ terms within the text body. After following these selection criteria, the numbers in the dataset $D$ dropped down to $305,418$ unique tweets identifiers and $304,885$ unique tweets.

The \texttt{geo.fullname} object was split into two subparts based on its [small region, larger region] data structure. This data structure was not the same for all the tweets in the dataset---some had single location detail such as just ``Melbourne", and ``New South Wales". In such cases, the single location details were considered small regions. Following this step, there were $3,724$ small region unique entries (shown in Table \ref{MegaGeoCOV-aus-small-location}) and $125$ larger region unique entries (shown in Table \ref{MegaGeoCOV-aus-larger-location})  in dataset $D$. As a general overview of the dataset, Table \ref{MegaGeoCOV-aus-locations} lists the top Australian locations (cities/towns/states) participating in the COVID-19 Twitter discourse, and Table \ref{MegaGeoCOV-aus-uigrams-bigrams} lists the most frequent unigrams and bigrams used by Australian Twitter users during the discourse.

\begin{table}[!htb]
\footnotesize
    \caption{Top Australian locations in MegaGeoCOV.}
    \label{MegaGeoCOV-aus-locations}
    \begin{subtable}{.5\linewidth}
      \centering
        \caption{Top small regions (N=3724)}
        \label{MegaGeoCOV-aus-small-location}
        \begin{tabular}{p{3.2cm}|p{2cm}}
        \hline
        \textbf{Small regions} & \textbf{Tweets}\\
        \hline
        Melbourne & 94,330\\
        Sydney & 70,118\\
        Brisbane & 21,298\\
        Perth & 18,143\\
        Adelaide & 13,372\\
        Canberra & 8,366\\
        Gold Coast & 6,483\\
        New Castle & 3,574\\
        Sunshine Coast & 2,843\\
        Central Coast & 2,190\\
        \hline
        \end{tabular}
    \end{subtable}%
    \begin{subtable}{.5\linewidth}
      \centering
        \caption{Top larger regions (N=125)}
        \label{MegaGeoCOV-aus-larger-location}
        \begin{tabular}{p{4.5cm}|p{1.5cm}}
        \hline
        \textbf{Larger regions} & \textbf{Tweets}\\
        \hline
        Victoria & 107,560\\
        New South Wales & 89,142\\
        Queensland & 37,107\\
        Western Australia & 20,259\\
        South Australia & 15,301\\
        Australia & 14,703\\
        Australian Capital Territory & 8,373\\
        Not Available & 4,678\\
        Tasmania & 3,280\\
        Northern Territory & 2,141\\
        \hline
        \end{tabular}
    \end{subtable}
\end{table}

\begin{table}[!htb]
\footnotesize
    \caption{20 most frequent unigrams and bigrams used by Australian Twitter users in the COVID-19 discourse.}
    \label{MegaGeoCOV-aus-uigrams-bigrams}
    \begin{subtable}{.47\linewidth}
      \centering
        \caption{Unigrams}
        \label{MegaGeoCOV-aus-uigrams}
        \begin{tabular}{p{2.7cm}|p{1.8cm}}
        \hline
        \textbf{Unigram} & \textbf{Frequency}\\
        \hline
        covid & 46,942\\
        lockdown & 34,016\\
        people & 30,936\\
        virus & 21,844\\
        vaccine & 19,380\\
        covid-19 & 18,132\\
        \#covid-19 & 17,231\\
        quarantine & 16,618\\
        pandemic & 15,842\\
        australia & 13,456\\
        mask & 12,936\\
        time & 12,891\\
        coronavirus & 12,602\\
        health & 12,111\\
        cases & 11,444\\
        \#coronavirus & 8,859\\
        government & 8,679\\
        nsw & 8,481\\
        home & 8,202\\
        work & 8,104\\
        \hline
        \end{tabular}
    \end{subtable}%
    \begin{subtable}{.53\linewidth}
      \centering
        \caption{Bigrams}
        \label{MegaGeoCOV-aus-bigrams}
        \begin{tabular}{p{4.2cm}|p{1.8cm}}
        \hline
        \textbf{Bigram} & \textbf{Frequency}\\
        \hline
        (`hotel', `quarantine') & 3,161\\
        (`wear', `mask') & 2,037\\
        (`2', `weeks')/('14', 'days')  & 1,974\\
        (`aged', `care') & 1,970\\
        (`wearing', `mask') & 1,517\\
        (`new', `cases') & 1,463\\
        (`social', `distancing') & 1,424\\
        (`public', `health') & 1,303\\
        (`new', `daily') & 1,302\\
        (`many', `people') & 1,239\\
        (`mental', `health') & 1,221\\
        (`federal', `government') & 1,154\\
        (`covid', `cases') & 1,098\\
        (`last', `year') & 1,077\\
        (`vaccine', `rollout') & 1,074\\
        (`stay', `home') & 1,037\\
        (`face', `mask') & 1,002\\
        (`tested', `positive') & 907\\
        (`covid', `vaccine') & 858\\
        (`covid', `test') & 805\\
        \hline
        \end{tabular}
    \end{subtable}
\end{table}

The dataset $D$ at this stage is $\{(t_{1},  tw_{1}, g_{1}), ... , (t_{N},  tw_{N}, g_{N})\}$, where $N=305,418$, the first component, $t_{1}, ... , t_{N}$, represents date/time attribute, the second component, $tw_{1}, ... , tw_{N}$, represents individual tweets, and the third component, $g_{1}, ... , g_{N}$, represents geolocation information of the individual tweets.

\subsection{Sentiment Analysis with BERT}
There exists a plethora of pre-trained sentiment analysis models and libraries suitable for sentiment analysis of short texts. Short-length texts and common use of informal grammar, abbreviations, spelling errors, and hashtags make it difficult in using pre-trained sentiment analyzers trained on formally written and typographical errors-free large-scale text corpora to handle sentiment analysis tasks on Twitter data. Further, in our case, we required a sentiment analyzer capable of understanding COVID-19 specific tweets.

Therefore, we finetuned a pre-trained language model, BERTweet \cite{nguyen2020bertweet}, for our sentiment analysis task. The language model has been reported to outperform existing state-of-the-art models across multiple NLP tasks including text classification. BERTweet has the same architecture as BERT$_{base}$ \cite{devlin2018bert} and is trained on 850 million English Tweets (cased) and additional 23 million COVID-19 English Tweets (cased) using the RoBERTa \cite{liu2019roberta} pre-training procedure. We finetuned the pre-trained BERTweet (bertweet-covid19-base-cased) model using the \texttt{transformers} library \cite{wolf2019huggingface} on the SemEval-2017 Task 4A dataset\footnote{https://alt.qcri.org/semeval2017/task4/} and achieved an accuracy of 0.7231 on the validation set built using the scikit-learn's train\_test\_split function \cite{pedregosa2011scikit} with parameters (given for results reproducibility) test\_size=0.2, random\_state=41, and stratify setting on the sentiment column. 

The fine-tuned model (hereafter termed as \textit{BERTsent}) outputs three labels each with a probability score for sentiment analysis: 0 representing ``negative" sentiment, 1 representing ``neutral" sentiment, and 2 representing ``positive" sentiment. The model effectively classifies sentences such as ``I had covid.” as negative and just the word “covid” as neutral by a significant probabilistic margin. We release both the PyTorch and TensorFlow versions of \textit{BERTsent} from the Hugging Face Hub\footnote{https://huggingface.co/rabindralamsal/BERTsent}.

Next, we used \textit{BERTsent} to compute sentiment probabilities for each tweet in dataset $D$. Dataset $D$ at this stage gets a new component, $sn_{1}, ... , sn_{N}$, that represents the sentiment of individual tweets. Output label with the highest probability was considered as the sentiment of a tweet. Dataset $D$ at this stage is:

$\{(t_{1},  tw_{1}, g_{1}, sn_{1}), ... , (t_{N},  tw_{N}, g_{N}, sn_{N})\}$.

\subsection{Topic Modelling}
\label{tp-model}
Next, we identify topics that best describe all the tweets in dataset $D$. We implemented one of the commonly preferred topic modelling techniques---Latent Dirichlet Allocation (LDA) \cite{blei2003latent}---using Gensim's LdaMallet module \cite{rehurek_lrec} which is a Python wrapper for LDA from MALLET \cite{mccallum2002mallet}. LDA maps all the tweets in dataset $D$ to the topics such that terms in each tweet are mostly captured by the topics. A ``topic" represents a group of words that often occur together. Algorithm \ref{lda} breifly summarizes the steps taken in implementing LDA on the tweets present in dataset $D$.

\begin{algorithm}
\caption{LDA implementation}
\label{lda}
\footnotesize

(i) $D_{LDA}$ $\gets$ all tweets present in $D$

(ii) Drop duplicate tweets from $D_{LDA}$

(iii) Clean tweets:

\hspace{7mm} (a) Ignore tweets with terms count $<$ 10

\hspace{7mm} (b) Transform tweets to lowercase

\hspace{7mm} (c) Remove URLs, mentions and consider only alphabets and digits

\hspace{7mm} (d) Remove extra spaces and paragraph breaks

(iv) Remove Stop words and tokenize each tweet into a list of words

(v) Identify frequently appearing bigrams in $D_{LDA}$ and add them to the list of words

(vi) Lemmatize each unigram present in the list of lists of words while considering only Noun part-of-speech

(vii) Construct word$<$-$>$integer\_ids mappings; design the bag-of-words format: list of (integer\_ids, token\_count) 2-tuples

(viii) Perform topic modelling

\For{each integer between 5 and 50 as number\_of\_topics}{
    Build LDA model
    
    Compute average topic coherence score based on the measure (c$_v$) proposed in \cite{roder2015exploring}
    }

(ix) Select the LDA model $M$ with highest average topic coherence score and human interpretability

(x) Use $M$ for assigning topics to tweets in dataset $D$
\end{algorithm}

Steps (iv), (v), (vii), and (viii) of Algorithm \ref{lda} were implemented using Gensim's Python library. For both unigrams and bigrams, the minimum term frequency was set to 500 to ignore sparsely appearing terms. For lemmatization, we used spaCy's Python library and considered only the Noun part-of-speech for building the topic models. Gensim's LdaMallet module was employed for building LDA models of a varying number of topics $k$. Having the ``right" $k$ solely based on mathematical goodness-of-fit does not necessarily mean that the topics have the best interpretability \cite{chang2009reading}. Therefore, the best $k$ was identified based on both the average topic coherence score and the human interpretability of the produced topics.

The value of $k$ was set in the range 5-50. LDA models were created for each $k$, and for each model the topic coherence scores were averaged. The highest average topic coherence scores were observed at $k=18$ and $k=22$; however, the interpretability of topics was relatively better at $k=18$. The final LDA model $M$ with $k=18$ was used for assigning topics to each tweet in Dataset $D$. \ref{lda-results} presents the LDA results obtained on $D_{LDA}$. Tweets were assigned topics based on a probability distribution generated by $M$---a tweet is assigned to a topic whose probability score is the highest in the distribution. 

With the addition of the topic component, $tp_{1}, ... , tp_{N}$ , dataset $D$ becomes:

$\{(t_{1},  tw_{1}, g_{1}, sn_{1}, tp_{1}), ... , (t_{N},  tw_{N}, g_{N}, sn_{N}, tp_{N})\}$

\subsection{Design of Time series}
\label{tsdesign}
Next, a time series dataset $D_{ts}$ was created based on dataset $D$ for the period January 1, 2020, to September 9, 2021. Dataset $D$ was grouped by the date/time component, $t_{1}, ... , t_{N}$, and the frequency of tweets across each day was summed for computing the volume of tweets over different topics and sentiments. From here, additional (number of topics=18 x number of sentiments=3) 54 components were generated, where each component represented topics and sentiments combined form.

$D_{ts}$ can be represented as a tensor of the following form: 

\begin{equation*}
\label{d_ts}
  D_{ts}: X_{tp^{j}sn^{k}}^{t^{i}}
\end{equation*}
where, index $i$ associates with the date component, index $j$ associates with the topic component, and index $k$ associates with the sentiment component. These indices take the values:  $i=618, ... , 1$ ($t^{618}$ representing January 1, 2020, and $t^{1}$ representing September 9, 2021); $j=0, ... , 17$; and $k=0,1,2$.

\subsubsection{Lagged time series}
For topic and sentiment components in $D_{ts}$, an additional of $l = 1, ... , 14$ days lagged components were generated to create a new time series dataset $D_{ts-lagged}$, that takes the following tensor form:

\begin{equation*}
\label{d_ts_lagged}
  D_{ts-lagged}: X_{(tp^{j}sn^{k})_{l}}^{t^{i}}
\end{equation*}

Generating the lagged components introduces NULL values in the last 14 samples of $D_{ts}$; therefore, $D_{ts-lagged}$ consists of tweets time series data for the period: January 15, 2020--September 9, 2021, after the loss of 14 days' data. $D_{ts-lagged}$ is created so that the forecasting models trained on it can regress on the lagged variables present in $D_{ts}$ to look up to 14 days back for making forecasts. The maximum lag of 14 was considered because of: (a) incubation period of the virus and suggested quarantine period \cite{pfizer}, (b) research works confirming the correlation between social media activities and future trends of the evolution of the virus \cite{li2020retrospective}.

\section{Experimentation and discussion}
\label{experimentation}

\subsection{Feature selection}
\label{featureselection}
At this stage, there are 54 components in $D_{ts}$. We performed feature selection based on Granger Causality \cite{granger1969investigating} to identify the set of features that are better predictors of daily confirmed COVID-19 cases. Tests were performed for all the variables in $D_{ts}$ to check if $X$ causes $y$, where $X = \{x_{1}, x_{2},...,x_{54}\}$, and $y$=COVID-19 confirmed cases. The data source for $y$ was \textit{OWID} \cite{owidcoronavirus}.

\textbf{Granger Causality} is a statistical concept that determines if a time series helps forecast another. A time series $x$ is said to``Granger-cause" a time series $y$ if the lagged values of $x$ contain information that helps predict $y$ exceeding the predictive ability carried by the lagged values of $y$ alone. 

\textit{Mathematical statement}: Granger causality supposes the following hypotheses---$H_0$ (null): $x$ does not Granger-cause $y$, $H_a$ (alternative): $x$ Granger-causes $y$. Both the time series need to be stationary i.e. parameters such as mean and variance should remain constant over time. To test $H_0$, the proper number of lags of $y$ to be included in an univariate autoregressive model of $y$ (Equation \ref{autoreg}) is identified using information criteria such as \textit{Akaike information criterion (AIC)} \cite{bozdogan1987model} and \textit{Bayesian information criterion (BIC) (also known as \textit{Schwarz Criterion})} \cite{schwarz1978estimating}. \textit{AIC} and \textit{BIC} are formally defined as:

\begin{equation}
    \label{aic}
    AIC = 2k - 2ln(\widehat{L})
\end{equation}

\begin{equation}
    \label{aic}
    BIC = ln(n)k - 2ln(\widehat{L})
\end{equation}
where $k$ is the number of estimated parameters (the variables in the model and the intercept), $\widehat{L}$ is a measure of model fit, and $n$ is the sample size.

We start with modelling an autoregressive model $y_{t}$ that has the lowest \textit{AIC} or \textit{BIC} value.

\begin{equation}
\label{autoreg}
    y_{t} =  a_{0}+a_{1}y_{t-1}+a_{2}y_{t-2}+...+a_{n}y_{t-n}+ e_{t}
\end{equation}

Next, the lagged values of $x$ are included into the model $y_{t}$.

\begin{equation}
    \label{autoregfull}
    y_{t} =  a_{0}+a_{1}y_{t-1}+a_{2}y_{t-2}+...+a_{n}y_{t-n}+ b_{s}x_{t-s}+...+b_{l}x_{t-l}+e_{t}
\end{equation}

The $s$ and $l$ parameters, in Equation \ref{autoregfull}, are the shortest and longest lag lengths for which the values of $x$ are significant. $H_0$ is accepted if and only if no lagged values of x are significant in Equation \ref{autoregfull}. The significance of the individual variables and their collective explanatory power is done based on t-test and F-test, respectively.  

\begin{figure}
  \includegraphics[width=\linewidth]{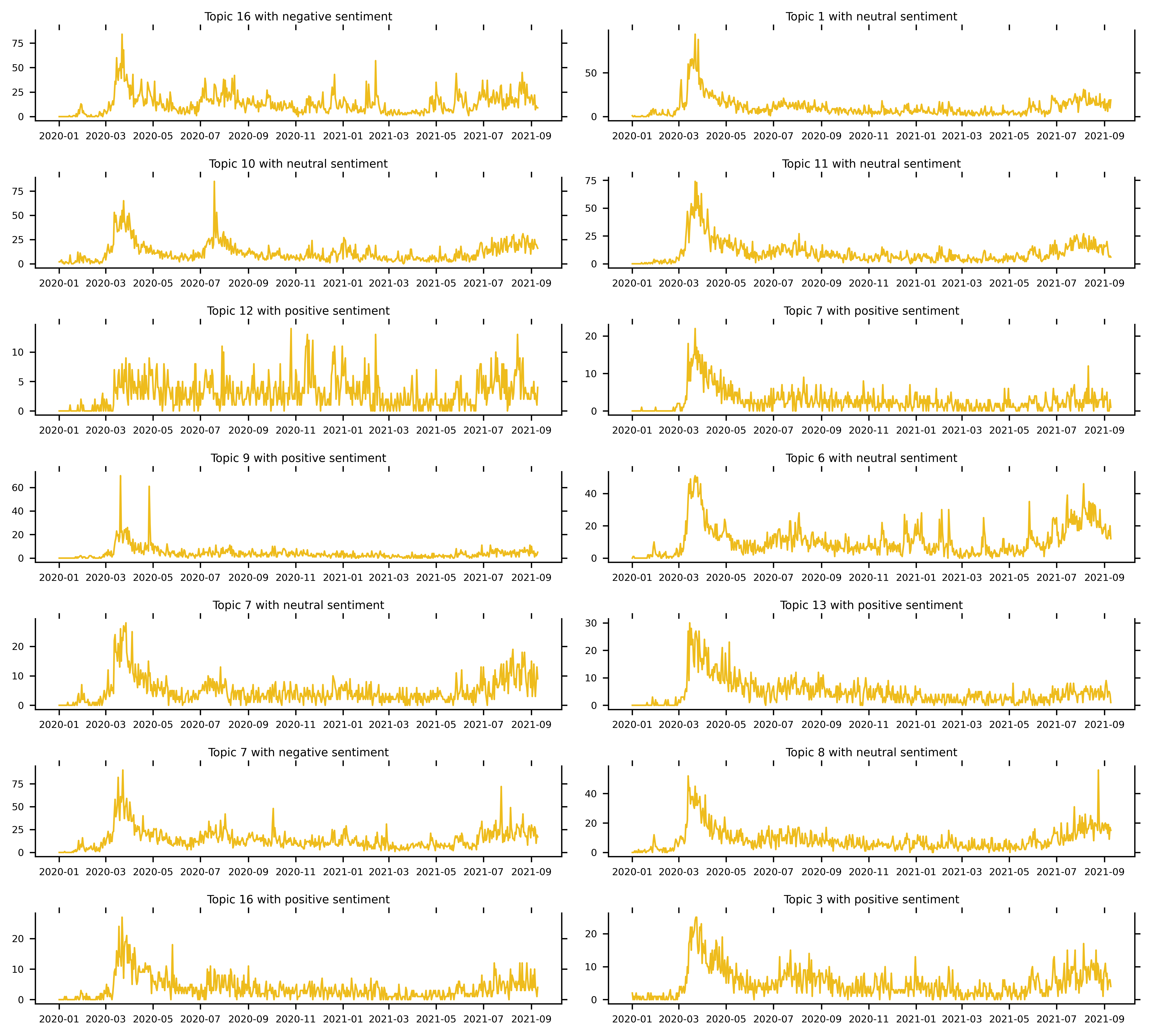}
  \caption{Plots of the variables (listed in Table \ref{gct-results}) in $D_{ts}$ that Granger-cause $y$ at most lags ($\geq$10). For each subplot, the vertical axis represents the \textit{count of tweets}, and the horizontal axis represents the \textit{date}.}
  \label{gct-results-plots}
\end{figure}

The causality test was performed between $y$ and each $x_i$ for the maximum lags of $14$ at 5\% significance level. We used Statsmodels' adfuller module \cite{seabold2010statsmodels} to implement the Augmented Dickey-Fuller (ADF) test \cite{dickey1979distribution} to check variables for stationarity. The test supposes the following hypotheses---$H_0$: Non Stationarity exists in the series, $H_a$: Stationarity exists in the series. Second-level differencing was required to make $y$ and all variables in $D_{ts}$ stationary. Table \ref{gct-results} lists the set of variables sorted based on the count of significant $p$-values i.e. count of lags at which a variable was observed Granger-causing $y$. The respective plots of these variables are shown in Figure \ref{gct-results-plots}.

\begin{table}[!h]
\footnotesize
\centering
\caption{Variables in $D_{ts}$ that Granger-cause $y$ at most lags (only $\geq$10 listed)}
\label{gct-results}
\begin{tabular}{l|p{2.5cm}|p{1.5cm}|l|p{2.5cm}|p{1.5cm}}
\hline
\textbf{variable} & \textbf{variable \newline definition} & \textbf{sig. \newline $p$-values} & \textbf{variable} & \textbf{variable \newline definition} & \textbf{sig. \newline $p$-values}\\
\hline
$X_{tp^{16}sn^{0}}^{t^{i}}$ & $topic_{16}$ negative & 14 & $X_{tp^{6}sn^{1}}^{t^{i}}$ & $topic_{6}$ neutral & 12\\
\hline
$X_{tp^{1}sn^{1}}^{t^{i}}$ & $topic_{1}$ neutral & 14 & $X_{tp^{7}sn^{1}}^{t^{i}}$ & $topic_{7}$ neutral & 12\\
\hline
$X_{tp^{10}sn^{1}}^{t^{i}}$ & $topic_{10}$ neutral & 14 & $X_{tp^{13}sn^{2}}^{t^{i}}$ & $topic_{13}$ positive & 12\\
\hline
$X_{tp^{11}sn^{1}}^{t^{i}}$ & $topic_{11}$ neutral & 14 & $X_{tp^{7}sn^{0}}^{t^{i}}$ & $topic_{7}$ negative & 11\\
\hline
$X_{tp^{12}sn^{2}}^{t^{i}}$ & $topic_{12}$ positive & 14 & $X_{tp^{8}sn^{1}}^{t^{i}}$ & $topic_{8}$ neutral & 11\\
\hline
$X_{tp^{7}sn^{2}}^{t^{i}}$ & $topic_{7}$ positive & 13 & $X_{tp^{16}sn^{2}}^{t^{i}}$ & $topic_{16}$ positive & 11\\
\hline
$X_{tp^{9}sn^{2}}^{t^{i}}$ & $topic_{9}$ positive & 13 & $X_{tp^{3}sn^{2}}^{t^{i}}$ & $topic_{3}$ positive & 10\\
\hline

\end{tabular}
\end{table}

\subsection{Forecasting models}
Autoregressive (\textit{AR}), Moving Average (\textit{MA}), \textit{ARMA}, Integrated \textit{ARMA} (\textit{ARIMA}), exogenous variables included \textit{ARIMA} (\textit{ARIMAX}), seasonal observations and errors-based (\textit{SARIMA}, \textit{SARIMAX}), \textit{Prophet}, Neural net-based (\textit{NeuralProphet}, \textit{Long Short-Term Memory}), and stochastic gradient boosting-based (\textit{XGBoost}) are some of the widely used time series forecasting models. Before getting started with experiments to address the research questions (\textit{RQ2}, \textit{RQ3}, and \textit{RQ4}) of this study, we fit the variable $y$ to multiple time series forecasting models for identifying the model that best explains the variable's trend. This way, going forward, it is justifiable to continue with the best model and introduce the social media context into the model. We used a machine learning python library, \textit{Auto TS}\footnote{https://github.com/AutoViML/Auto\_TS}, for building multiple traditional-based, FB Prophet, and XGBoost models on $y$ and identified the best model based on the reported Root Mean Square Error (RMSE) (Equation \ref{rmse}) scores. The training and testing were performed using the expanded window cross-validation (using the library's default parameters). Table \ref{autotsresults} shows the results provided by Auto TS.

\begin{table}[!h]
\footnotesize
\centering
\caption{Best forecasting model for $y$}
\label{autotsresults}
\begin{tabular}{@{}p{8.5cm}|l@{}}
\hline
\textbf{approach} & \textbf{Avg. RMSE}\\
\hline
traditional model$^a$ (ARIMA of $p=1$,$d=1$,$q=3$)$^b$ & 135.387\\
additive model (FB Prophet) & 236.427\\
machine learning model (XGBoost) & 341.8\\
\hline
\end{tabular}
\\
$^a$also involves the participation of the models such as \textit{AR}, \textit{MA}, \textit{ARIMA}, \textit{SARIMA}. $^b$the traditional models and their mathematical structures are discussed later in Section \ref{arimax-models}.
\end{table}

We also did experiments with neural network models; the results were not encouraging; maybe the amount of data (this study uses 618 days' of data) is not sufficient to fully exploit the forecasting capabilities of neural-based models. The results, reported in Table \ref{autotsresults}, suggest that the traditional models significantly explain the cases trend compared to the additive approach-based FB Prophet and the gradient boosting-based XGBoost model. From here, to address the research questions \textit{RQ2}, \textit{RQ3} and \textit{RQ4}, the design of forecasting models is done in two phases. First, we design ARIMA with exogenous variables (ARIMAX) models to show that the inclusion of social media data provides additional forecasting capabilities. Second, we design Vector Autoregressive (VAR) models to forecast the number of COVID-19 cases, seven days into the future, using the same set of variables.

\subsubsection{ARIMAX models}
\label{arimax-models}
\textbf{Mathematical definition.} Given a time series $y_{t}$, the autoregressive part, AR$(p)$, can be defined as:

\begin{equation}
\label{ar}
    y_{t} = \beta + \epsilon_{t} + \sum\limits_{i=1}^p \theta_{i} y_{t-i}
\end{equation}
where, $\beta$ is a constant, $\epsilon_{t}$ is the error at time $t$, and $p$ is the number of lags of the prior values of $y_{t}$ to be considered for regression.

Equation \ref{ar} can be made more concise (shown in Equation \ref{con-ar}) by introducing the back-shift operator (a.k.a. lag operator) L, as $L^{n} y_{t} = y_{t-n}$. 

\begin{equation}
    \label{con-ar}
    y_{t} = \Theta(L)^{p} y_{t} + \epsilon_{t}
\end{equation}
where, $\Theta(L)^{p}$ is the polynomial function of $L$ of order $p$.

Similarly, for the same time series $y_{t}$, the moving average part, MA$(q)$ can be defined as:

\begin{equation}
    \label{ma}
    y_{t} = \Phi(L)^{q} \epsilon_{t} + \epsilon_{t}
\end{equation}
where, $q$ is the number of lags of the prior values of error to be considered for regression, and $\Phi$ is defined similar to $\Theta$.

The sum of AR$(p) $ and MA$(q) $ models forms the ARMA$(p,q)$ model, which is defined as:

\begin{equation}
    \label{arma}
    y_{t} = \Theta(L)^{p} y_{t} + \Phi(L)^{q} \epsilon_{t} + \epsilon_{t}
\end{equation}

Further, to deal with non-stationary time series, an integration operator $\Delta^{d}$ is introduced and defined as: $y_{t}^{[d]} =\Delta^{d} y_{t} = y_{t}^{[d-1]} - y_{t-1}^{[d-1]}$, where $d$ is the order of differencing required to make the non-stationary time series stationary. When an ARMA$(p,q)$ model is fitted on the integrated time series, the model is termed as ARIMA$(p,d,q)$ and represented as:

\begin{equation}
    \label{arima}
    \Delta^{d} y_{t} = \Theta(L)^{p} \Delta^{d} y_{t} + \Phi(L)^{q} \Delta^{d} \epsilon_{t} + \Delta^{d} \epsilon_{t}
\end{equation}

\begin{equation}
    \label{con-arima}
    \Theta(L)^{p} \Delta^{d} y_{t} = \Phi(L)^{q} \Delta^{d} \epsilon_{t}
\end{equation}

When the ARIMA$(p,d,q)$ models consider exogenous variables into account, the models are termed ARIMAX$(p,d,q)$ models and represented as:

\begin{equation}
    \label{arimax}
    \Theta(L)^{p} \Delta^{d} y_{t} = \Phi(L)^{q} \Delta^{d} \epsilon_{t} + \sum_{i=1}^{n} \beta_{i} x^{i}_{t}
\end{equation}
where, $n$ is the number of exogenous variables $x^{i}_{t}$ with $\beta_{i}$ as their respective coefficients.

Exogenous variables at time $t$ are the independent variables that influence the dependent variable at $t$. ARIMAX models do not regress on the lagged values of such variables; instead, they are computed outside the system and used for predicting the dependent variable. In our case, the social media variables are the exogenous ones; however, our designed lagged time series dataset $D_{ts-lagged}$ also incorporates lagged values so that the time series models can look back up to 14 days and make forecasts accordingly.

We use Root Mean Square Error (RMSE), Mean Absolute Percentage Error (MAPE), and Coefficient of Determination (R2) as the measures for assessing the quality of predictions made by the forecasting models. For $N$ number of observations with $x_{i}$ representing true values and $\widehat{x}_{i}$ representing predicted values, RMSE, MAPE, and R2 are mathematically defined as:

\begin{equation}
    \label{rmse}
    RMSE = \sqrt{\frac{\sum_{i=1}^{N}(x_{i}-\widehat{x}_{i})^{2}}{N}}
\end{equation}

\begin{equation}
    \label{mape}
    MAPE = \frac{100}{N} \sum_{i=1}^{N}\left | \frac{x_{i}-\widehat{x}_{i}}{x_{i}} \right |
\end{equation}

\begin{equation}
\label{R2}
R2 = 1 - {\sum_i (x_i - \widehat{x}_i)^2\over \sum_i (x_i - \bar{x})^2}
\end{equation}

We fit ARIMA$(p,d,q)$ models on y, and ARIMAX$(p,d,q)$ on $y$ and the variables (alongside their lags available through $D_{ts-lagged}$) in $D_{ts}$ that Granger-cause $y$ at all 14 lags. We mark the ARIMA$(p,d,q)$ models as baseline model candidates and the ARIMAX$(p,d,q)$ models as social media model candidates. All the models were fitted on the data observed up to August 26, 2021, and tested on the data observed between August 27, 2021, and September 9, 2021. The best fit was determined based on the reported AIC scores---lower the AIC, better the fit. The results from the training are shown in Table \ref{fit-results} for both set of models.

\begin{table}[!htb]
\footnotesize
    \caption{Results from training. Models are ranked based on their AIC scores.}
    \label{fit-results}
    \begin{subtable}{.5\linewidth}
      \centering
        \caption{Top 5 baseline models}
        \begin{tabular}{l|l|l}
        \hline
        \textbf{(p,d,q)} & \textbf{AIC} & \textbf{RMSE}\\
        \hline
        (6, 2, 7) &	6118.50 & 37.78\\
        \hline
        (5, 2, 8) & 6118.80 & 37.81\\
        \hline
        (7, 2, 5) & 6119.06 & 37.89\\
        \hline
        (7, 2, 8) & 6120.12 & 37.70\\
        \hline
        (7, 2, 6) & 6120.21 & 37.87\\
        \hline
        \end{tabular}
    \end{subtable}%
    \begin{subtable}{.5\linewidth}
      \centering
        \caption{Top 5 social media models}
        \begin{tabular}{l|l|l}
        \hline
        \textbf{(p,d,q)} & \textbf{AIC} & \textbf{RMSE}\\
        \hline
        (2, 2, 3) &	5941.08 & 32.97\\
        \hline
        (1, 2, 4) &	5942.43 & 33.05\\
        \hline
        (2, 2, 2) &	5945.95 & 33.26\\
        \hline
        (4, 2, 3) & 5957.88 & 33.12\\
        \hline
        (4, 2, 2) & 5960.05 & 33.41\\
        \hline
        \end{tabular}
    \end{subtable}
\end{table}

\begin{figure}[t]
\begin{subfigure}{.70\textwidth}
  \centering
  \includegraphics[width=1\linewidth]{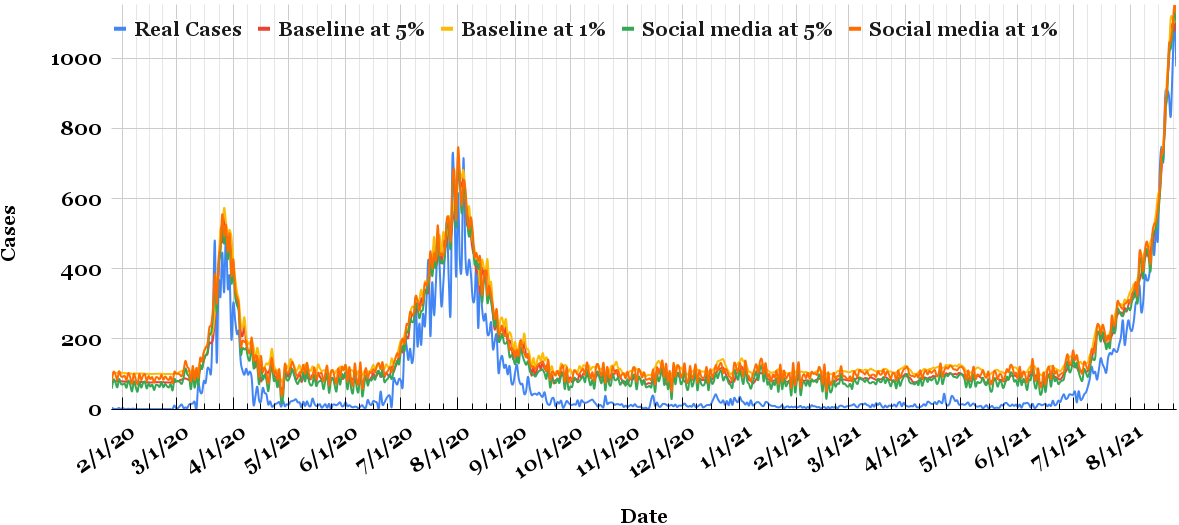}
 \caption{Results on train data}
  \label{papers-dist-type}
\end{subfigure}%
\begin{subfigure}{.30\textwidth}
  \centering
  \includegraphics[width=0.6\linewidth]{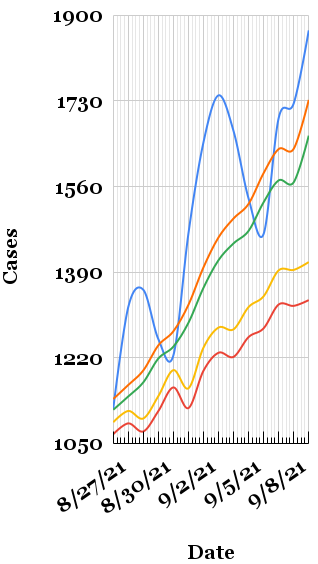}
 \caption{Results on test data}
  \label{papers-dist-year-type}
\end{subfigure}
\caption{COVID-19 confirmed cases versus the cases predicted by the baseline and social media models at 1\% and 5\% significance levels.}
\label{models-results}
\end{figure}

Since all social media models report lower RMSE on the training data compared to the baseline models, it is evident that the inclusion of the social media variables for modelling does help explain the dependent variable better (12.73\% improvement over the best baseline model) compared to using just the lagged values of the dependent variable. It is also apparent that the best-fitted social media model requires lower lag parameters for both autoregressive and moving-average processes compared to the best-fitted baseline model. For forecasting the daily COVID-19 cases for the test period, we selected the ARIMA$(6,2,7)$ model as the baseline model and the ARIMAX$(2,2,3)$ as the social media model. The residuals from both models were further checked for the presence of any possible patterns. For both models, the residual correlograms showed autocorrelations near-zero (insignificant) for all lags. Table \ref{results} presents the forecasting results obtained using baseline and social media models at 1\% and 5\% significance, and Figure \ref{models-results} plots the forecasts of the models from both training and testing phases.

\begin{table}[!h]
\footnotesize
\centering
\caption{Results (upper values) from test data. Baseline model versus Social media model at 1\% and 5\% significance.}
\label{results}
\begin{tabular}{l|p{1cm}|l|l|l|l}
\hline
\multirow{2}{*}{\textbf{date}} & \multirow{2}{*}{\textbf{cases}} & \multicolumn{2}{c|}{\textbf{baseline}} & \multicolumn{2}{c}{\textbf{social media}}\\
& & \textbf{at 5\%} & \textbf{at 1\%} & \textbf{at 5\%} & \textbf{at 1\%}\\
\hline
\textbf{2021-08-27} & 1119 & 1068 & 1092 & 1116 & 1138\\
\hline
\textbf{2021-08-28} & 1321 & 1090 & 1114 & 1143 & 1166\\
\hline
\textbf{2021-08-29} &1355 & 1074 & 1099 & 1171 & 1195\\
\hline
\textbf{2021-08-30} & 1257 & 1114 & 1144 & 1219 & 1244\\
\hline
\textbf{2021-08-31} & 1225 & 1161 & 1195 & 1242 & 1272\\
\hline
\textbf{2021-09-01} & 1467 & 1120 & 1159 & 1289 & 1325\\
\hline
\textbf{2021-09-02} & 1648 & 1194 & 1240 & 1358 & 1399\\
\hline
\textbf{2021-09-03} & 1741 & 1230 & 1280 & 1413 & 1459\\
\hline
\textbf{2021-09-04} & 1670 & 1221 & 1276 & 1447 & 1496\\
\hline
\textbf{2021-09-05} & 1536 & 1261 & 1320 & 1472 & 1525\\
\hline
\textbf{2021-09-06} & 1466 & 1279 & 1342 & 1529 & 1586\\
\hline
\textbf{2021-09-07} & 1696 & 1326 & 1393 & 1572 & 1634\\
\hline
\textbf{2021-09-08} & 1725 & 1323 & 1394 & 1568 & 1635\\
\hline
\textbf{2021-09-09} & 1870 & 1334 & 1410 & 1661 & 1731\\
\hline
\multicolumn{2}{c|}{\textbf{RMSE}} & 342.58 & 295.68 & 175.31 & 143.76\\
\hline
\multicolumn{2}{c|}{\textbf{MAPE}} & 19.36\% & 16.29\% & 9.24\% & 7.61\%\\
\hline
\multicolumn{2}{c|}{\textbf{R2}} & 0.67 & 0.68 & 0.75 & 0.75\\
\hline

\end{tabular}
\end{table}

On the testing data, the social media models introduce 48.83\% and 51.38\% improvements on RMSE over the baseline models at 5\% and 1\% significance, respectively. These significant improvements confirm that the social media discourse indeed is a good predictor for pandemic-related forecasting models. In Table \ref{results}, if we look at the data observed after September 1, 2021, the forecasting ability of the baseline model begins to be off by significant margins, while the social media model seems to be catching up with the trend of the everyday cases with small errors.

The testing timeline in this study, a steep-hill curve (also shown in Figure \ref{models-results}), was the most suitable region (compared to monotonically ascending regions) for examining the effect of exogenous variables that might influence the variable to be forecasted. Based on the results presented in this section, we conclude that the latent variables extracted from the COVID-19 specific social media discourse can be good predictors of the pandemic's daily cases, and these variables are predictive of the steep-hill curve of COVID-19 cases during an ongoing wave.

Continuing with the idea that the social media variables are predictive of our dependent variable, in the next section, we fit VAR models to forecast the COVID-19 cases in Australia for the next 7 days.

\subsubsection{VARMA models}
Vector Autoregressive Moving-Average (VARMA) models are multivariate linear time series models generally used for simultaneous modeling of multiple stationary time series and generating simultaneous forecasts of the independent variables in the system. Mathematically, a VARMA$(p,q)$ model is defined as:

\begin{equation}
    \label{varma}
    y_t = c + \sum_{j=1}^{p} \Theta_j y_{t-j} + \sum_{k=1}^{q} \Phi_k \epsilon_{t-k} + \epsilon_t
\end{equation}
where, $y_t$ is an $n\times1$ vector of distinct dependent time series variables at $t$, $c$ is an $n\times1$ vector of constant in each equation, $\Theta_j$ is an $n\times n$ matrix of autoregressive coefficients, $\Phi_k$ is an $n\times n$ matrix of moving-average coefficients and $\epsilon_t$ is an $n\times1$ vector of error terms.

From our experiments, we observed that the inclusion of the moving-average part of the VARMA$(p,q)$ models did not improve the quality of the forecasts compared to using the autoregressive part alone. Therefore, we considered VAR$(p)$ models (defined as Equation \ref{var}) for our multivariate time series forecasting.

\begin{equation}
    \label{var}
    y_t = c + \sum_{j=1}^{p} \Theta_j y_{t-j} + \epsilon_t
\end{equation}

We fitted multiple VAR$(p)$ models, where $0\leq p \leq 20$, on the variables (except for the lagged ones) used by the social media model in Section \ref{arimax-models} to forecast the COVID-19 cases for the next 7 days. The results from the VAR order selection and the forecasts made by the best fitted VAR model are shown in Table \ref{var-models} and Figure \ref{forecasts}, respectively. We observed the lowest AIC score with the VAR$(15)$ model. The social media model from Section \ref{arimax-models} had the autoregressive process of lag order of 2, implying that looking back up to 15 days best describes our dependent variable---we had the lagged time series dataset $D_{ts-lagged}$ designed in such as way that the lag order of 1 included the past 14 days' data, the lag order of 2 included the past 15 days' data, and so on. We observe the same mathematical implication here from the best-fitted VAR model.

\bigskip
\begin{minipage}[c]{0.40\textwidth}
\footnotesize
    \centering
    \captionof{table}{VAR order selection---fitting VAR models on $D_{ts}$. Lowest AIC score is highlighted.}
    \label{var-models}
    \begin{tabular}{c|c}
    \hline
      \textbf{parameter \textit{p}} & \textbf{AIC}\\
      \hline
      0	& 28.60\\
      1	& 23.33\\
      2 & 22.90\\
      3	& 22.69\\
      ... & ...\\
      15 & \textbf{22.50}\\
      16 & 22.52\\
      \end{tabular}
    \end{minipage} 
    \hfill
  \begin{minipage}[c]{0.55\textwidth}
    \centering
    \includegraphics[width=1\textwidth]{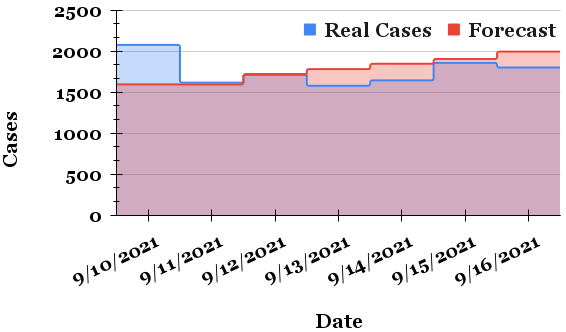}
    \captionof{figure}{Forecast of COVID-19 cases for the next 7 days with VAR$(15)$ model. $MAPE = 9.08\%$ (overall); $MAPE=6.74\%$ (excluding the 9/10/2021's sudden rise).}
    \label{forecasts}
  \end{minipage}
\bigskip

The VAR$(15)$ model was used for forecasting the COVID-19 cases in Australia one week in advance from September 10, 2021, to September 16, 2021. The forecasts and the deviations from the actual cases are illustrated in Figure \ref{forecasts}. The RMSE and the MAPE of the overall forecasts were 224.65 and 9.08\%, respectively. Excluding the September 10, 2021's sudden rise, the model reported RMSE of 142.8 and MAPE of 6.74\%. Out of the 7 days' forecasts, the model forecasted the cases almost perfectly for 3 days and with small margin of errors for the other 3 days. The VAR model can be deployed for making forecasts using unseen tweets. Its dependency is on dataset $D_{ts}$, which is based on the outputs generated by BERTsent and the LDA-based topic model. After a collection of a statistically significant number of social media conversations related to an event, similar topic model can be trained and used along side BERTsent to generate a time series dataset identical to $D_{ts}$ as discussed in Section \ref{tsdesign}.

\subsection{Comparison with the existing studies}
In this study, we proposed a representation for microblog conversations that can represent the volume of social media activity (conversations) feature at a more granular level to decrease the intensity of possible forecast biases. In the existing literature, the ``volume” feature includes social media search indexes, category-based counts, and overall count strategies. Use of the ``volume" feature keeps computational complexity to minimal as we maintain only the counts of tweets based on a strategy. Notably, such models can be deployed on small-scale infrastructures. However, those models get heavily affected by avalanches of auto-generated conversations. Therefore, this study proposed a representation for microblog conversations to break the ``volume” feature to more granular levels in order to decrease the dependency of the models on one or a few thematic counts.

From Table~\ref{autotsresults}, it is evident that the traditional forecasting models significantly explain the trend of the daily confirmed COVID-19 cases in Australia compared to additive-based, machine learning, and neural models. This observation is in agreement with what has been reported in earlier studies \cite{papastefanopoulos2020covid,saba2021machine} that involved the forecast of COVID-19 cases. Moving on, in this section, we compare the forecasting ability of our social media model with existing studies that use social media ``volume" feature for designing discourse-based forecasting models. To compare our methodology (identifying relevant exogenous variables through latent variables search), we fit various volumetric features considered by existing studies, as exogenous variables to forecast our dependent variable $y$.

\textbf{Social media-based volumetric features.} The following volumetric features were considered as exogenous variables for comparison against the variables identified by our latent variables search methodology.

\textbf{(i) Search indexes}: Google Trends\footnote{https://trends.google.com/trends/?geo=AU} was considered the data source for social media search indexes. The platform provides the popularity of search queries on Google across various geographical regions. The popularity of a search query is provided through a set of numbers (between 0-100) for each day, where the peak value ``100" is the highest point on the graph for the given region and timeline. The platform gives the daily search interests for a search query only for a timeline of 9 months at most; beyond that range, week-level search interests are provided. For this study, we extracted search interests in three different blocks (search trend blocks) for the period January 1, 2020, and September 9, 2021, for the following terms: \texttt{dry cough}, \texttt{chest distress}, \texttt{coronavirus}, \texttt{fever}, and \texttt{pneumonia}. The search trend blocks were created with overlaps to scale the second and third blocks relative to the first. The daily search interests in the second and third blocks were re-scaled by the blocks' respective scaling factors as:

\begin{equation}
\label{factor}
current~scale~value * factor = previous~scale~value
\end{equation}

\begin{figure}
     \centering
     \begin{subfigure}[b]{0.5\textwidth}
         \centering
         \includegraphics[width=\textwidth]{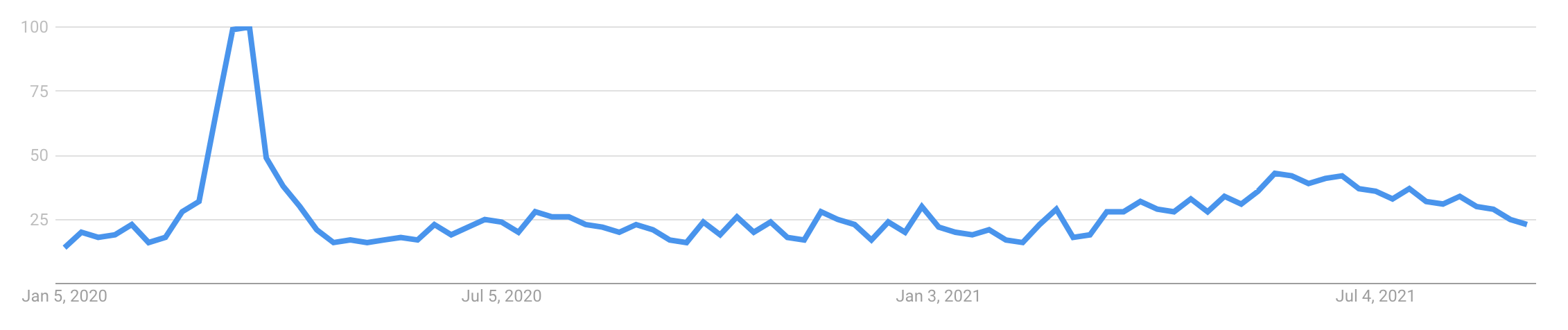}
         \caption{Search interest for ``dry cough" }
         \label{fig:drycough}
     \end{subfigure}
     \begin{subfigure}[b]{0.48\textwidth}
         \centering
         \includegraphics[width=\textwidth]{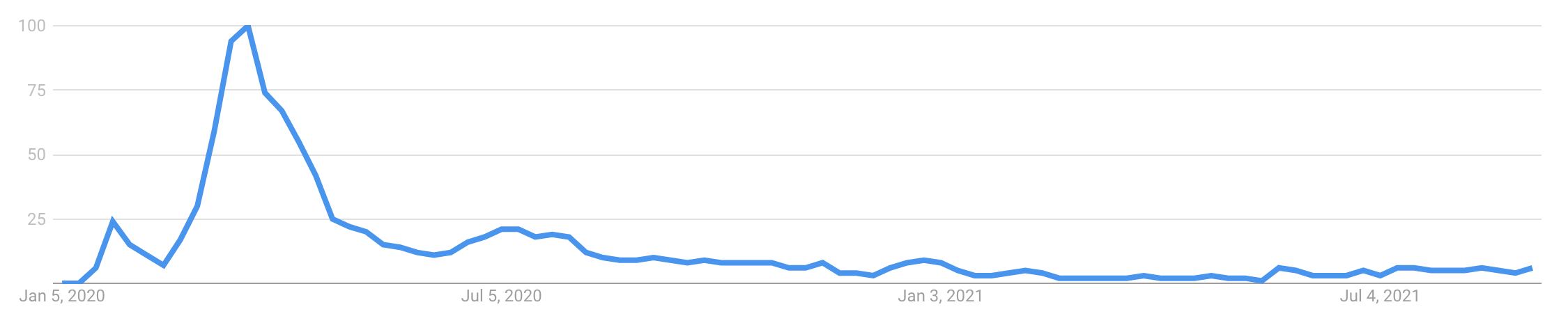}
         \caption{Search interest for ``coronavirus" }
         \label{fig:coronavirus}
     \end{subfigure}

     \begin{subfigure}[b]{0.5\textwidth}
         \centering
         \includegraphics[width=\textwidth]{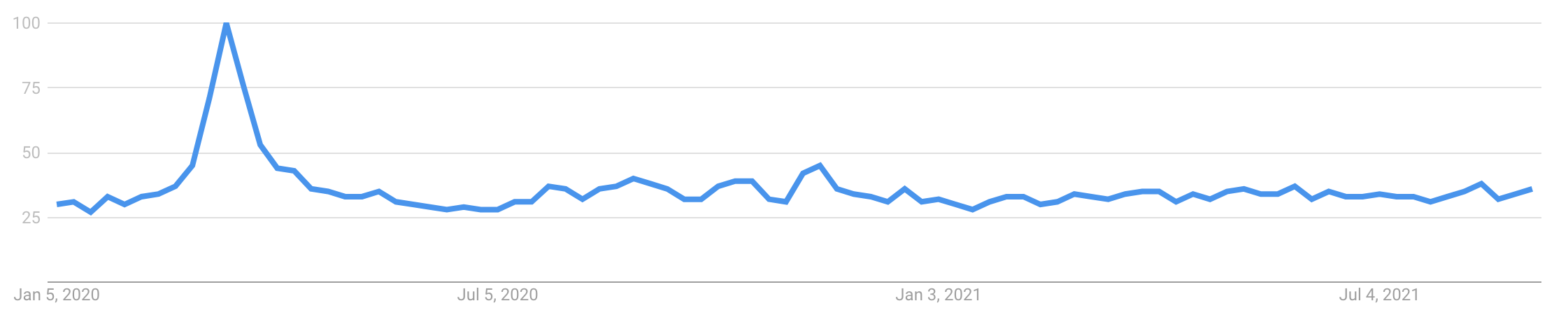}
         \caption{Search interest for ``fever"}
         \label{fig:fever}
     \end{subfigure}
     \begin{subfigure}[b]{0.48\textwidth}
         \centering
         \includegraphics[width=\textwidth]{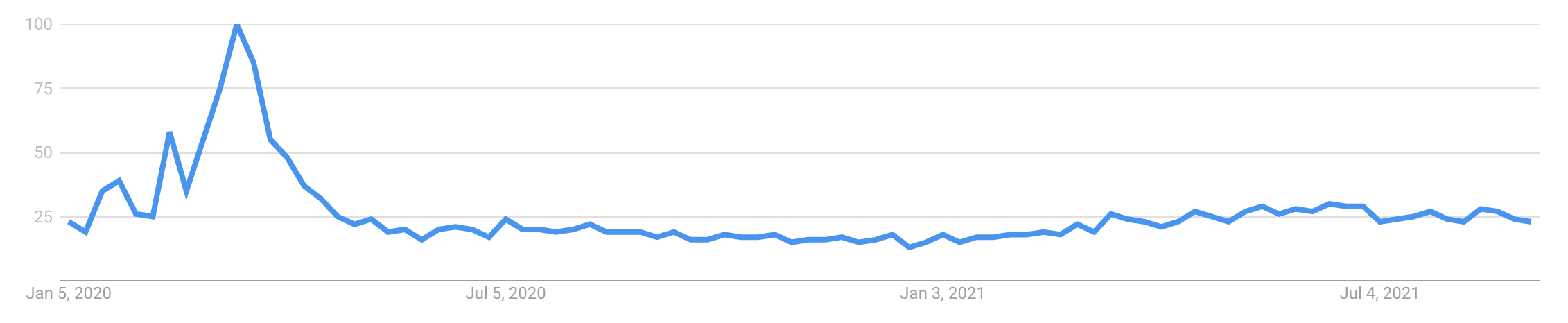}
         \caption{Search interest for ``pneumonia"}
         \label{fig:pneumonia}
     \end{subfigure}

     \begin{subfigure}[b]{\textwidth}
         \centering
         \includegraphics[width=\textwidth]{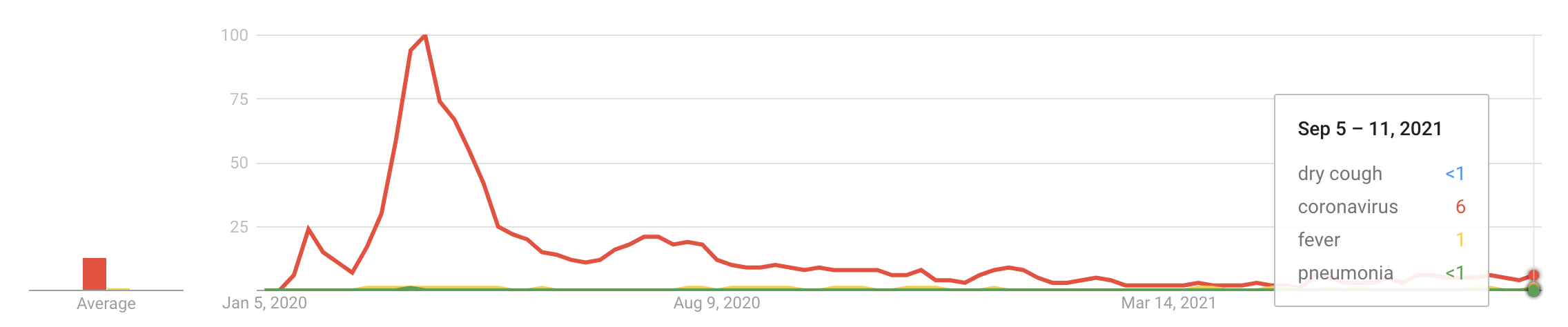}
         \caption{Search interests for all four terms (relative)}
         \label{fig:all:search}
     \end{subfigure}
     
        \caption{Search interests data retrieved from Google Trends for the period January 1, 2020, and September 9, 2021.}
        \label{search_trends}
\end{figure}

Figure \ref{search_trends} plots the daily Google search interests for the search terms. The term ``chest distress" was excluded since it did not have significant search interest in Australia. Figure \ref{fig:all:search} is the plot for all search terms relative to each other. It is evident from the plot that the search interests for the term ``coronavirus" was significantly higher compared to other terms.

\textbf{(ii) Sick posts}: We processed all the Twitter conversations in dataset $D$ through the LDA model designed in Section \ref{tp-model} to create ``sick" related posts' time series. Tweets with the highest score in the probability distribution for topic ``6" were considered as ``sick" related posts. Some salient words in topic ``6" include (sorted based on the influence) \texttt{test}, \texttt{case}, \texttt{testing}, \texttt{isolation}, \texttt{symptom}, \texttt{clinic}, \texttt{lab}, \texttt{isolate}, \texttt{swab}, \texttt{fever}, \texttt{throat}, \texttt{trace}, \texttt{temperature}, \texttt{quarantine}, \texttt{positive}, \texttt{tracer}, \texttt{carrier}, \texttt{diagnosis}, \texttt{pathology}, \texttt{vitamin}.

\textbf{(iii) Overall posts}: A daily distribution was maintained for the Twitter conversations present in dataset $D$ to create the ``overall" posts' time series.

Next, we created additional 14 lagged variables for each time series to assist the models to look back up to 14 days for making forecasts (dataset $D_{ts-lagged}$ followed the same implementation). Table \ref{comp-sota} and \ref{res-train} summarize the results from fitting ARIMAX models on different sets of exogenous variables considered in the existing studies. We use the same training and testing timeline as the social media model designed in Section \ref{arimax-models}.

\begin{table}[h!]
\footnotesize
\centering
\caption{Comparison of our latent variables search methodology with existing studies that use social media-based volumetric features.}
\label{comp-sota}
\begin{tabular}{p{4.2cm}|l|l|p{0.8cm}|l|l|p{0.9cm}} 
\hline
                               & \multicolumn{3}{c|}{\textbf{at 5\%}} & \multicolumn{3}{c}{\textbf{at 1\%}}  \\ 
\hline
                               & \textbf{RMSE} & \textbf{MAPE} & \textbf{R2}            & \textbf{RMSE} & \textbf{MAPE} & \textbf{R2}             \\ 
\hline
Baseline$^a$                       &  342.58    &  19.36\%    &     0.67          &  295.68    &  16.29\%    &      0.68          \\ 
\hline
Search index (dry cough)$^b$       &   326.93   &  17.76\%    &        0.68       &  277.22    &   14.74\%   &       0.7         \\ 
\hline
Search index (coronavirus)$^b$     &   307.48   &  16.98\%    &        0.7       &   258.716   &   13.75\%   &        0.7        \\ 
\hline
Search index (fever)$^b$           &   344.15   &   19.49\%   &       0.67        &   297.28   &   16.4\%   &          0.68      \\ 
\hline
Search index (pneumonia)$^b$       &   266.13   &  14.51\%    &      0.67         &   223.15   &   11.79\%   &        0.68        \\ 
\hline
Search indexes Combined$^b$        &  241.23    &   13.1\%   &       0.66        &   200.40   &   10.62\%   &       0.67         \\ 
\hline
Sick posts$^c$                   &  283.44    &   15.71\%   &       0.68        &   239.68   &   12.72\%   &       0.69         \\ 
\hline
Sick posts + Search indexes combined & 198.52 & 10.29\% & 0.7 & 160.62 & 8.52\% & 0.70\\
\hline
Overall posts$^d$                  &  289.16    &   16.07\%   &       0.73        &   241.44   &  12.84\%    &        0.73        \\ 
\hline
Latent variables search$^e$ &   175.31   &   9.24\%   &    0.75           &   143.76   &   7.61\%   &       0.75         \\
\hline
\end{tabular}
$^a$fitted solely on $y$. Exogenous variables: $^b$\cite{qin2020prediction, li2020retrospective, cousins2020regional}, $^c$\cite{shen2020using}, $^d$\cite{comito2021covid}. $^e$this study.
\end{table}

\begin{table}[h!]
\footnotesize
\centering
\caption{Results from fitting the exogenous variables listed in Table \ref{comp-sota} and their respective 14 days' lags  against 84 weeks of data (January 15, 2020, to August 26, 2021).}
\label{res-train}
\begin{tabular}{p{4.2cm}|p{2.5cm}|p{2.5cm}|l|l} 
\hline
                           & \textbf{Best fitted model} & \textbf{Exo. Variables count} & \textbf{AIC} & \textbf{RMSE}  \\ 
\hline
Baseline                   &      ARIMA(6,2,7) &    -        &  6118.50   &   37.78    \\ 
\hline
Search index (dry cough)   &       ARIMAX(9,2,9)&     1 and its 14 lags      &  6019.93   &    37.46   \\ 
\hline
Search index (coronavirus) &       ARIMAX(7,2,5) &     1 and its 14 lags      &  6013.5   &    37.51   \\ 
\hline
Search index (fever)       &       ARIMAX(5,2,8) &     1 and its 14 lags      &   5993.47  &   37.55    \\ 
\hline
Search index (pneumonia)   &       ARIMAX(6,2,9) &     1 and its 14 lags      &  6001.28   &    37.52   \\ 
\hline
Search indexes Combined    &     ARIMAX(7,2,8)	 &     4 and respective 14 lags     &  6085.15   &    36.53   \\ 
\hline
Sick posts                 &      ARIMAX(8,2,7)  &     1 and its 14 lags    &  5989.78   &   37.12    \\
\hline
Sick posts + Search indexes combined & ARIMAX(3,2,9)& 5 and respective 14 lags & 6069.28 & 35.77\\
\hline
Overall posts              &       ARIMAX(4,2,5) &     1 and its 14 lags      &  5991.94   &   37.34    \\ 
\hline
Latent variables search   &    ARIMAX(2,2,3) &     14 and respective 14 lags      &  5941.08   &   32.97    \\
\hline
\end{tabular}
\end{table}

Table \ref{comp-sota} reports the RMSE, MAPE, and R2, of the baseline model, existing studies, and this study at both 5\% and 1\% significance. The results show that our methodology outperforms the existing studies that use social media-based volumetric features to forecast the daily confirmed COVID-19 cases. Except for the search term \texttt{fever}, the search interests of the other three terms included in the experimentation, i.e., \texttt{dry cough}, \texttt{coronavirus}, and \texttt{pneumonia}, seem to provide additional forecasting abilities (compared to the baseline model that was regressed only on $y$). When all search terms were combined and fitted, there were further improvements observed in both RMSE and MAPE. The best-fitted model for the ``sick" related posts performed poorly compared to the search indexes combined model. We performed an additional modeling by combining and fitting the exogenous variables associated with sick posts and all search indexes, and observed significant improvements in RMSE and MAPE; the metrics improved to 198.52 and 10.29\% at 5\%, and 160.62 and 8.52\% at 1\%. The overall posts model performed on par with the sick posts model, providing evidence that the count strategy, be it category-based or general, offers limited forecast capability. Overall, our latent variables search methodology achieves the lowest RMSE and MAPE and the highest R2 at both significant levels.

To demonstrate the robustness of our methodology, in Table \ref{res-train}, we provide the results (p, d, and q parameters of the best-fitted models, their respective exogenous variables counts, and AIC/RMSE scores) obtained while fitting the exogenous variables listed in Table 12 and their respective 14 days’ lags against 84 weeks of data, i.e., January 15, 2020, to August 26, 2021. The results show that the exogenous variables identified by our latent variables search methodology explain the dependent variable better compared to the existing studies in the literature. For the 84 weeks of data, our social media model benchmark the lowest RMSE of 32.97 and is followed by the Sick posts + Search indexes combined model with an RMSE of 35.77. All the models with exogenous variables achieved better RMSE scores than the ARIMA-based baseline model.

\textbf{Issue with search interests.} Search interests are ``broad" in nature---a search for ``coronavirus" can relate to multiple use cases, such as checking top stories, querying updates and local information, and accessing health information (symptoms, prevention, treatments). Search interests do not provide the granular-level distinction of the use case unless the search terms are more specific, such as ``melbourne covid hotspots today", ``coronavirus symptoms", and ``covid hotline melbourne". Therefore, while designing interpretable forecasting models it is critical to exploit the public conversations for searching latent variables that carry granular-level details regarding an event. Besides, services such as Google Trends can retire, or data extraction can be made limited as the platforms upgrade to different versions. However, discourse-based models entirely rely on the conversations and can have applications outside of Twitter-verse.

\subsection{The research questions}
In this section, we address the four research questions (RQ1--4) that this study sets out to answer.

Modeling of Twitter data for region-specific analyses requires a large amount of geotagged tweets. For addressing \textit{RQ1}, we curated a large-scale geotagged tweets dataset---\textit{MegaGeoCOV}---targetting the public COVID-19 discourse. We used Twitter’s Academic Track-based Full-archive search and count APIs to access the numbers presented in Table \ref{MegaGeoCOV}. Between January 01, 2020, and September 9, 2021, the minimum number of tweets (for the specific set of keywords and hashtags mentioned in Section \ref{datacollection}) was 59.6k and the maximum was 25.8 million, with a mean of 4.62 million. Among those numbers, the volume of geotagged tweets were observed between 0.449\%--1.43\%. Although the geotagged volume is considerably limited, the experiments from this study suggest that ``what is currently available" is satisfactory for designing similar discourse-based forecasting models. We addressed \textit{RQ2} by performing Granger causality tests on the time series that were created based on the geotagged Twitter data. We observed the presence of latent variables within the data that Granger-caused the daily COVID-19 confirmed cases time series. Some such variables (granger-causing at lags $\geq 10$ out of 14 lags) are listed in Table \ref{gct-results}. The methodology for the identification of such variables is discussed in Section \ref{featureselection}. We also observed that the identified Granger-causing latent variables provide additional prediction capability to time series forecasting models (this observation addresses \textit{RQ3}). We noticed that the inclusion of social media variables for modeling introduced 12.73\% improvement on the training data, and above 48\% improvements (at 1\% and 5\% significance) on the testing data over the baseline model (discussed in Section \ref{arimax-models}). Furthermore, ``the volume of public discourse in the last few days" being predictive of the steep-hill curve of COVID-19 cases during an ongoing wave address our \textit{RQ4}. The latent variables (variables in $D_{ts}$) are the outputs of every day's tweet volume. The forecasts produced by the ARIMAX and VAR models designed in this study verify that the volume of public discourse is predictive of the COVID-19 cases' steep-hill trend.

\section{Conclusion}
\label{conclusion}
In this paper, a sentiment-involved topic-based latent variables search methodology was proposed for time series analysis of publicly available COVID-19 related Twitter conversations. A language model trained on 850 million English Tweets (cased) and additional 23 million COVID-19 English Tweets (cased) using the RoBERTa pre-training procedure was finetuned for the sentiment analysis task, and LDA was performed for identifying the hidden topics within the conversations. The proposed methodology was implemented on the COVID-19 cases in Australia and Twitter conversations generated within the country between January 1, 2020, and September 9, 2021. ARIMA models (baseline model candidates) were fitted on $y$=\textit{Australian COVID-19 cases}, and ARIMAX models (social media model candidates) were fitted on $y$ and the social media variables (alongside their lags) that Granger-cause $y$ the most (at all 14 lags). Experimental results from the training showed that the inclusion of social media variables for modeling brings in 12.73\% improvement over the baseline model compared to using just the lagged values of $y$. While, on the testing data, the social media models introduced 48.83\% and 51.38\% improvements on RMSE over the baseline models at 5\% and 1\% significance, respectively. Considering the same set of variables used by the social media model, a VAR model was used for forecasting the COVID-19 cases in Australia one week in advance from September 10, 2021, to September 16, 2021. Out of the 7 forecasts, the model predicted the cases almost perfectly for 3 days and with small margin of errors for the other 3 days---with RMSE and MAPE ranging between 142.8--224.65 and 6.74--9.08\%, respectively (the upper values of these metrics is outcome of the September 10, 2021's sudden rise in the cases with respect to the next 6 days).

This study confirms the presence of a relationship between latent social media variables and COVID-19 daily cases. The literature seems to have overlooked the social media perspective of the COVID-19 time series analyses. The inclusion of social media variables alongside native epidemiological data (causes, risk factors, population descriptors, etc.) can be beneficial for an early forecast of an epidemic/pandemic's future courses. One of the limitations of this study is that only the social media variables are included in the time series analysis. Social media, and especially microblogging platforms, are more skewed towards tech-savvy users and younger populations. Further, the study considers only three categories of sentiment---positive, neutral, and negative---and does not consider other possible categories such as hate, offense, and irony. Furthermore, filtration of misinformative tweets can be an additional tweets selection procedure before time series are constructed. These limitations could be an important research avenue while designing next generations of discourse-based forecasting models.

\section*{CRediT authorship contribution statement}
\textbf{RL} performed Conceptualization, Data curation, Methodology, Software, Visualization, Writing--first draft. \textbf{AH} and \textbf{MRR} performed Conceptualization, Supervision, Writing--Review \& Editing.

\section*{Acknowledgements} This study was supported by the \textit{Melbourne Research Scholarship} from the University of Melbourne, Australia. We (the authors) are thankful to the \textit{Nectar Research Cloud} for providing us with a large-volume compute instance (32 VCPUs, 288GB memory, 20TB volume) running NeCTAR Ubuntu 21.04 (Hirsute). We are also grateful to \textit{DigitalOcean} for supporting us with a cloud-based VM during the data curation phase of this study as a part of the COVID-19 relief initiative ``Hub for Good". We also express our special gratitude to \textit{Twitter} for their continuous support towards the social media research community---the wide range of currently available public APIs from the platform and the latest academic track initiative are commendable.

\section*{Declaration of competing interest}
The authors declare that they have no known competing financial interests or personal relationships that could have appeared to influence the work reported in this paper.

\newpage 
\appendix

\section{LDA results on $D_{LDA}$}
\label{lda-results}
{\tiny \begin{longtable}{c|p{12cm}}
\hline
\textbf{Topic} & \textbf{Salient words}\\
\hline
\multirow{3}{*}{0} & lockdown, pm, rule, idea, message, panic, detail, move, meeting, announcement, restriction, location, gathering, situation, detention, notice, prime\_minister, mess, looking\_forward, regulation \\
\hline
\multirow{3}{*}{1} & food, order, book, shop, market, price, water, store, delivery, supermarket, restaurant, paper, supply, stock, demand, sale, stuff, trade, cafe, list, product, customer, shortage, grocery\\
\hline
\multirow{3}{*}{2} &   family, friend, hope, love, mate, house, shit, thought, heart, member, visit, guy, daughter, wife, mother, girl, party, partner, son, movie, dad, anxiety, brother, memory, sister, colleague, loved\_one, neighbour, kind, hug, spirit, song, prayer, soul, sunshine\\
\hline
\multirow{3}{*}{3} &   time, thing, moment, ship, air, lung, fire, cruise, trip, passenger, winter, crew, plane, hell, summer, pain, island, tip, weather, get\_back, spring, ruby\_princess, hope, quality, doubt, trouble, board, tour, track, smoke, breathe, omg, port, storm, boat\\
\hline
\multirow{3}{*}{4} &   year, team, game, event, show, season, video, player, sport, club, tv, challenge, watch, fan, race, art, music, crowd, training, play, session, ground, tennis, football, ticket, court, venue, ball, goal, episode, win, series, cricket, artist, film, star, host, horse, content, performance, league, competition, song, entertainment, gig\\
\hline
\multirow{3}{*}{5} &   death, people, number, rate, infection, risk, population, datum, transmission, protest, freedom, idiot, conspiracy, spread, exposure, control, theory, toll, site, suicide, factor, stat, mortality, prevent, evidence, confirmed\_case, statistic, count, analysis, victim, every\_day, protester, survivor, fatality, cases\_death, surge\\
\hline
\multirow{3}{*}{6} &   day, today, test, case, person, hour, testing, isolation, yesterday, symptom, contact, tomorrow, area, period, week, clinic, morning, site, line, act, last\_week, lab, drive, delay, queue, household, isolate, swab, fever, throat, trace, temperature, quarantine, positive, tracer, carrier, diagnosis, pathology, caution, vitamin\\
\hline
\multirow{3}{*}{7} &   people, world, country, life, rest, war, leader, million, threat, earth, around\_world, citizen, pressure, stop, die, moron, stupidity, worry, shit, covidiot, problem, spanish\_flu, kill, narrative, planet, prison, mentalhealth, years\_ago, faith, enemy, weapon, danger, livelihood, estate, liberty, bullet, fighting, destruction, frustration\\
\hline
\multirow{3}{*}{8} & health, issue, system, advice, problem, expert, science, effect, research, safety, emergency, condition, treatment, mental\_health, concern, evidence, disease, management, trial, effort, scientist, solution, trust, officer, report, process, authority, drug, brain, damage\\
\hline
\multirow{3}{*}{9} & news, media, story, fact, article, election, app, information, fear, comment, answer, view, truth, tweet, info, twitter, vote, source, journalist, ad, opinion, page, claim, president, statement\\
\hline
\multirow{3}{*}{10} & mask, hospital, hand, patient, doctor, ace, staff, care, nurse, centre, phone, distance, shopping, icu, eye, ppe, pace, practice, bed, work, line, nose, capacity, folk, guideline, mouth, limit, nursing, lady\\
\hline
\multirow{3}{*}{11} & work, job, business, worker, support, service, money, company, industry, cost, office, pay, healthcare, access, economy, leave, loss, bill, payment, driver, bus, sector, university, frontline, income, tax, \\
\hline
\multirow{3}{*}{12} & state, case, border, vic, nsw, travel, restriction, outbreak, premier, flight, new\_case, control, record, sa, victorian, wave, gladys, hotel\_quarantine, cluster, traveller, arrival, closure, bubble, community\_transmission, update, ban, region, hotspot, territory, exemption\\
\hline
\multirow{3}{*}{13} & community, response, measure, part, change, decision, level, impact, economy, nation, point, situation, law, recovery, strategy, opportunity, crisis, sense, society, step, term, history, experience, reality, role, behaviour, contract, thread, lock, model\\
\hline
\multirow{3}{*}{14} & vaccine, flu, vaccination, pfizer, group, risk, age, jab, dose, study, delta, choice, blood, chance, strain, shot, astrazeneca, variant, type, appointment, get\_vaccine, protection, immunity, pfizer\_vaccine, reaction, virus, clot, target, gp, supply\\
\hline
\multirow{3}{*}{15} & lockdown, week, month, melbourne, sydney, end, city, weekend, beach, stage, road, street, exercise, adelaide, town, restriction, suburb, stayhome, start, curfew, first\_time, half, last\_year, staysafe, apartment, rock, melb, sight, stage\_lockdown\\
\hline
\multirow{3}{*}{16} & quarantine, home, hotel, police, student, security, parent, facility, place, hotel\_quarantine, room, care, program, staff, purpose, force, airport, guard, adult, member, breach, station, protocol, inquiry, two\_week, standard, fine, requirement\\
\hline
\multirow{3}{*}{17} & government, auspol, morrison, plan, australian, govt, failure, policy, leadership, power, responsibility, labor, disaster, leader, action, blame, federal\_government, politician, deal, attack, excuse, crisis, insider, lack, lie, climate, minister, vaccine\_rollout, credit, recession \\
\hline
\end{longtable}}

\newpage
\bibliographystyle{elsarticle-num} 
\bibliography{ref}

\end{document}